# Mamba meets crack segmentation

Zhili He[a], Yu-Hsing Wang[a*]

[a]*Department of Civil and Environmental Engineering,*

*The Hong Kong University of Science and Technology, HKSAR, China*

**ABSTRACT:**

Cracks pose safety risks to infrastructure that cannot be overlooked and serve as a vital indicator of structural performance deterioration. Enhancing the precision and efficiency of crack segmentation is a prominent research topic. The prevailing architectures in existing crack segmentation networks predominantly consist of convolutional neural networks (CNNs) or Transformers. Nonetheless, CNNs exhibit a deficiency in global modeling capability, hindering their representation to entire crack features. Transformers can capture long-range dependencies but suffer from high and quadratic complexity. Recently, the emergence of a novel architecture, Mamba, has garnered considerable attention due to its linear spatial and computational complexity, as well as its powerful global perception capability. This study explores the representation capabilities of Mamba to crack features. Specifically, this paper uncovers the intrinsic connection between Mamba and the attention mechanism, providing a profound insight—an attention perspective—into interpreting Mamba and devising a novel Mamba module following the principles governing attention blocks, namely CrackMamba. We compare CrackMamba with the most prominent visual Mamba modules, Vim and Vmamba, on two datasets comprising asphalt pavement and concrete pavement cracks, and steel cracks, respectively. The quantitative results demonstrate that CrackMamba stands out as the sole Mamba block consistently enhancing the baseline model's performance across all evaluation measures, while reducing its parameters and computational costs. Moreover, this paper substantiates that Mamba can achieve global receptive fields through both theoretical analysis and visual interpretability. The discoveries of this study offer a dual contribution. First, as a plug-and-play and simple yet effective Mamba module, CrackMamba exhibits immense potential for integration into various crack segmentation models. Second, the proposed innovative Mamba design concept, integrating Mamba with the attention mechanism, holds significant reference value for all Mamba-based computer vision models, not limited to crack segmentation networks, as investigated in this study.

## 1. Introduction

Defects inevitably occur, accumulate, and propagate during the service life of civil infrastructure due to the intricate working environment. These defects not only pose significant safety hazards to structures but also threaten their service life. Among the various forms of structural defects, cracks are a primary manifestation and a pivotal indicator of structural performance degradation, demanding meticulous attention [1].

Routine crack inspection serves as an effective and widely employed method for mastering crack information of infrastructure. Presently, the predominant mode of inspection is manual visual examination conducted onsite, a practice that is both time-consuming and labor-intensive. Moreover, the subjective nature of human assessment introduces a notable level of uncertainty and inconsistency in the inspection outcomes [2]. Recognizing these

* Corresponding author.

*E-mail addresses:* ceyhwang@ust.hk (Y.-H. Wang).

limitations, there is a growing interest among researchers and practitioners in integrating cutting-edge artificial intelligence (AI) and computer vision (CV) technologies into crack inspection practices to achieve intelligent, efficient, and precise crack identification [3][4][5]. These AI-driven crack identification methodologies are typically categorized into three distinct groups based on the nature of the researched CV tasks: image-level crack classification, box-level crack detection, and pixel-level crack segmentation. Notably, He et al. [2] have identified crack segmentation as the most appropriate and accurate approach for representing crack features, thus, the research focus of this paper is crack segmentation.

Initially, crack segmentation models were constructed using convolutional neural networks (CNNs) [1][6][7]. However, considering that on the one hand the ability of CNNs to capture long-distance dependencies is so limited due to the inherent locality of convolutions, and on the other hand, crack pixels are typically dispersed globally on crack images in their length directions, it is challenging for CNN-based crack segmentation networks to comprehensively model the entirety of crack features. In contrast, a newer architecture than CNNs, Transformers, based on self-attention [8][9], inherently excels in modeling long-term context information. Consequently, Transformers have been integrated into crack segmentation networks either to replace CNNs [10] or to work with them [2], bringing the networks global perceptual capabilities. The frameworks incorporating Transformers [2][10] have demonstrated the benefits of effectively modeling long-range context features in crack segmentation tasks, further underscoring the necessity for crack segmentation networks to possess global modeling capabilities. Despite the advantages, the computational and spatial complexity of Transformers grows quadratically with the length of input sequences, presenting an efficiency bottleneck. As a result, researchers have endeavored to refine the architecture of Transformer to reduce computational and spatial costs without compromising its global perception abilities [11][12]. Various efficient modules have been proposed, such as Sparse Transformers [13], Linear attention [14], and FlashAttention [15]. However, these advanced modules only optimize the architecture of the vanilla Transformer without fundamentally innovating its core structure, namely, the scaled dot product self-attention. Given the well-established maturity and widely proven generalization and robustness capabilities of the vanilla Transformer, researchers and engineers still tend to favor the conventional Transformer architecture. The newer and more intricate Transformer structures, despite their novelty, have not yet emerged as the mainstream choice due to this preference.

Recently, in the realm of one-dimensional (1D) sequence modeling, an architectural innovation known as Mamba [16], coming from state space models (SSMs) [17][18] rather than the above-mentioned self-attention, has garnered significant attention because Mamba's powerful global modeling capabilities, state-of-the-art (SOTA) performance across diverse tasks encompassing multiple modalities, such as natural language processing (NLP), audio modeling, and genomics modeling, and more importantly, its linear complexity. Notably, pioneering work by Zhu et al. [19] and Liu et al. [20] has introduced Mamba into the domain of computer vision, resulting in the development of the Vision Mamba (Vim) model [19] and the VMamba model [20], respectively. These studies have demonstrated the conspicuous success of Mamba in learning general visual representation and global features of images. Specifically, Vim and VMamba have not only surpassed many well-established vision networks in performance but have also exhibited faster processing speeds and lower memory and computing resource requirements. Building on these achievements, a range of well-crafted models based on Vim or VMamba have been effectively deployed in specific downstream tasks of CV, such as Mamba-UNet [21] for medical image analysis and RS3Mamba [22] for remote sensing image segmentation. The series of developments underscore Mamba's versatility and immense potential to supplant CNNs and Transformers to be the next generation foundational architecture of CV. However, despite these advancements, the application of Mamba in the context of crack images remains unexplored. This poses an intriguing question regarding Mamba's ability to effectively represent and comprehend crack features. Addressing this question is of paramount importance, as successful utilization of Mamba in crack data analysis could revolutionize the field of crack identification, propelling it into a new stage of research and innovation.

Therefore, this study investigates the representation ability of Mamba to crack data. Initially, a classical and influential crack segmentation model, CrackSeU [1], is selected, along with two representative and publicly available crack datasets encompassing cracks from distinct domains. The reason of choosing the datasets is to examine if Mamba can consistently improve performance of the model to the cracks from different domains. The first dataset, Deepcrack [1][41], encompasses asphalt pavement cracks and concrete pavement cracks, while the second dataset, Steelcrack [2], comprises cracks in steel structures. Subsequently, Vim and VMamba are integrated into CrackSeU, and the enhanced networks are trained and evaluated on the two crack datasets. Unfortunately, the experimental outcomes in Subsection 5.1 reveal that Vim and VMamba cannot consistently enhance and, in some cases, even diminish the model's performance. Upon thorough investigation, it is discovered that the essence of Mamba and the attention mechanism [1][23] is unified in dynamic parameters, therefore, Mamba can be viewed as a type of attention. Given the universal success of the attention mechanism in crack identification [1][23][24][25], this study redesign the visual Mamba modules to propose the CrackMamba block, which aligns with the attention modules. Notably, the experimental results in Subsection 5.1 demonstrate that CrackMamba stands out as the sole Mamba module capable of consistently enhancing the crack recognition performance of CrackSeU across both datasets. Furthermore, CrackSeU combined with CrackMamba (CrackSeU + CrackMamba) requires fewer parameters and lower computational resources compared to the standard CrackSeU model. From the insights gained through this research, several key conclusions can be drawn: (1) the performance of Mamba modules is highly related to the tasks, and Vim and VMamba blocks are proved unsuitable for crack segmentation tasks; (2) the structural composition of Mamba blocks is crucial and significantly influences the performance of Mamba-based AI models, therefore, it is necessary to design the structures of the Mamba modules tailored to specific tasks; (3) the innovative perspective introduced, known as the attention perspective, serves as a valuable guide in shaping Mamba blocks and also deepens the comprehension of Mamba; and (4) it is believed that the proposed Mamba design concept, that designing Mamba modules aligning with attention modules, holds enlightening significance for a wide array of CV tasks which either have employed or are considering the utilization of visual Mamba models, not limited to the crack identification problem, as explored in this study. Moreover, this paper demonstrates that Mamba can bring global receptive fields from both theoretical and visual interpretability standpoints.

To summarize, the principal contributions of this research are outlined as follows:

(1) Mamba and visual Mamba, including Vim and VMamba, are introduced into the domain of crack segmentation. Detailed illustration and analysis, often overlooked in existing literature related to Mamba, are provided in the appendices to ensure clarity for the civil engineering audience.

(2) This research offers a heightened and profound perspective for comprehending Mamba, termed the attention perspective. Subsequently, a novel, straightforward, yet highly effective Mamba module, named CrackMamba, is designed following the principles of attention modules.

(3) Vim, VMamba, and CrackMamba are evaluated on two representative crack datasets, with the findings indicating that CrackMamba is the only Mamba block to consistently enhance crack segmentation performance. Additionally, CrackMamba reduces the number of parameters and computational complexity of the baseline model. These outcomes not only validate the effectiveness of CrackMamba and emphasize its substantial application potential in crack segmentation models, but also confirm the success of the proposed attention perspective and highlight the importance of designing Mamba blocks that incorporate the principles of the attention mechanism, which offers significant reference value for various Mamba-based CV models, not solely limited to the crack segmentation frameworks investigated by this study.

(4) This paper demonstrates the global modeling capability of Mamba through both the theoretical analysis and visual interpretability perspectives.

The structure of this paper is outlined as follows: Section 2 offers detailed descriptions of Mamba, visual Mamba, and CrackMamba. Section 3 presents the adopted segmentation model, the integration of Mamba and the segmentation network, and the crack datasets. Implementation details, encompassing the training policy and the evaluation measures, are thoroughly discussed in Section 4. Section 5 delves into the quantitative experimental results and provides an in-depth examination of the global receptive field of Mamba. Finally, Section 6 concludes the study and offers suggestions for future research endeavors. The implementation code for CrackMamba and a tutorial on installing Mamba are openly accessible at https://github.com/hzlbbfrog/CrackMamba.

## 2. Mamba

**Notations.** Let us introduce the notation rules adopted in this paper to avoid confusion. Following [26], matrices or tensors in neural network are denoted by non-tilt boldface uppercase letters, such as $\mathbf{A}$ and $\mathbf{X}$. Vectors are denoted by non-tilt boldface lowercase letters, such as $\mathbf{x}$ and $\mathbf{u}$. Scalars are represented by tilt non-bold lower-case letters, such as $x$ and $t$. Additionally, vectors are all column vectors. For instance, $\mathbf{x} = [x_0, x_1, \dots, x_{L-1}]^\top$ is adopted to represent the vector $\mathbf{x}$ formed by $L$ scalars.

### *2.1. Vanilla Mamba*

The evolution of state space models (SSMs) unfolds through two pivotal iterations: Structured state space sequence models, also known as Structured SSMs or S4 models [17][18], progress into Selective SSMs, referred to as SSMs+Selection or S6 models (aka Mamba) [16]. These iterations are elaborated upon in the subsequent contents.

The basic idea of S4 models comes from the state space equations in the modern control theory domain, and they are adopted to describe a linear time invariance (LTI) system:

$$\dot{\mathbf{x}}(t) = \mathbf{A}\mathbf{x}(t) + \mathbf{B}\mathbf{u}(t), \tag{1a}$$

$$\mathbf{y}(t) = \mathbf{C}\mathbf{x}(t) + \mathbf{D}\mathbf{u}(t), \tag{1b}$$

For a first-order system, the parameters in the above equations are scalars. Here, we only consider second-order or higher-order systems. Hence, $\mathbf{x}(t)$ represents the state vector of the system, $\mathbf{u}(t)$ is the system input, and $\mathbf{y}(t)$ denotes the system output. $\mathbf{A}$, $\mathbf{B}$, $\mathbf{C}$, and $\mathbf{D}$ are the state matrices used to describe the system. Since this system is a time invariance system, $\mathbf{A}$, $\mathbf{B}$, $\mathbf{C}$, and $\mathbf{D}$ are all constant matrices for any time $t$.

SSMs mainly make three modifications to the above state space equations. First, SSMs simplify $(1b)$ and omit $\mathbf{D}$ in $(1b)$ (or equivalently, $\mathbf{D} = \mathbf{0}$) [18]. This simplification is reasonable because $\mathbf{D}\mathbf{u}(t)$ can be viewed as a shortcut connection in neural networks and can be computed easily. Second, the input and output are assumed to be the 1D continuous signals. Third, the notations are slightly changed. Thus, the state space equations in the domain of SSMs can be defined as

$$\dot{\mathbf{h}}(t) = \mathbf{A}\mathbf{h}(t) + \mathbf{B}x(t), \tag{2a}$$

$$y(t) = \mathbf{C}\mathbf{h}(t), \tag{2b}$$

where the hidden state variable is changed to $\mathbf{h}(t) \in \mathbb{R}^{N\times 1}$, and $N$ denotes the dimension of the state vector. The system input is $x(t)$, and $y(t)$ still denotes the output, with $t$ representing the continuous time index. Clearly, $\mathbf{A} \in \mathbb{R}^{N\times N}$, $\mathbf{B} \in \mathbb{R}^{N\times 1}$, and $\mathbf{C} \in \mathbb{R}^{1\times N}$ based on the matrix operation rule. For the subsequent content, the symbols' meanings are consistent with Eqs. $(2a)$ and $(2b)$ unless specified otherwise. It is worth noting that the research core of all SSMs is how to effectively represent Eqs. $(2a)$ and $(2b)$ and effectively implement them in AI models and computing hardware.

S4 models first discretize the above state space equations since the calculations in the neural work are always

discrete:

$$\mathbf{h}_k = \overline{\mathbf{A}}\mathbf{h}_{k-1} + \overline{\mathbf{B}}x_k, \tag{3a}$$
$$y_k = \overline{\mathbf{C}}\mathbf{h}_k, \tag{3b}$$

At this time, Eqs. $(3a)$ and $(3b)$ describe the mapping between two discrete signals (or sequences): $\mathbf{x} \in \mathbb{R}^{L\times 1} = [x_0, x_1, \dots, x_{L-1}]^\top \Rightarrow \mathbf{y} \in \mathbb{R}^{L\times 1} = [y_0, y_1, \dots, y_{L-1}]^\top$, where $L$ denotes the length of the sequences. In the above equations, $k \in \{0,1,\dots, L-1\}$ denotes the discrete index, and $\overline{\mathbf{A}} \in \mathbb{R}^{N\times N}$, $\overline{\mathbf{B}} \in \mathbb{R}^{N\times 1}$, and $\overline{\mathbf{C}} \in \mathbb{R}^{1\times N}$ are the discrete parameters corresponding to the continuous parameters $\mathbf{A}$, $\mathbf{B}$ and $\mathbf{C}$, and they can typically be expressed as

$$\overline{\mathbf{A}} = f_{\mathbf{A}}(\Delta, \mathbf{A}), \overline{\mathbf{B}} = f_{\mathbf{B}}(\Delta, \mathbf{A}, \mathbf{B}), \text{and } \overline{\mathbf{C}} = \mathbf{C}. \tag{4}$$

The most straightforward discretization method is the Euler discretization method (see Appendix A.1), however, it is not precise enough. Thus, S4 models adopt a more advanced method, the bilinear discretization method, where $\overline{\mathbf{A}}$, $\overline{\mathbf{B}}$, and $\overline{\mathbf{C}}$ can be further formulated as

$$\overline{\mathbf{A}} = (\mathbf{I} - \Delta/2 \cdot \mathbf{A})^{-1}(\mathbf{I} + \Delta/2 \cdot \mathbf{A}), \tag{5}$$
$$\overline{\mathbf{B}} = (\mathbf{I} - \Delta/2 \cdot \mathbf{A})^{-1}\Delta\mathbf{B}, \tag{6}$$

and

$$\overline{\mathbf{C}} = \mathbf{C}, \tag{7}$$

where $\Delta \in \mathbb{R}^+$ represents the time step size of discretization, and $\mathbf{I} \in \{0,1\}^{N\times N}$ is the identity matrix. The proof of the bilinear discretization method is provided in Appendix A.2. Second, S4 models leverage the high-order polynomial projection operator (HiPPO) Matrix to initialize $\mathbf{A}$ to handle long-range dependencies, and S4 models find that in contrast to random initialization, the HiPPO Matrix initialization can improve the performance on the sequential MNIST classification benchmark from $60\%$ to $98\%$ [17][18]. HiPPO Matrix can be defined by the following formula:

$$\text{HiPPO Matrix} \quad \mathbf{A}_{ij} = -\begin{cases} (2i+1)^{1/2}(2j+1)^{1/2} & \text{if } i > j \\ i+1 & \text{if } i = j \, . \, i,j \in [0,1,\dots, N-1]. \\ 0 & \text{if } i < j \end{cases} \tag{8}$$

Third, S4 models combine Eqs. $(3a)$ and $(3b)$ for the entire sequence:

$$\mathbf{y} = \overline{\mathbf{K}} \circledast \mathbf{x}, \tag{9a}$$
$$\overline{\mathbf{K}} = \left(\overline{\mathbf{C}}\overline{\mathbf{B}}, \overline{\mathbf{C}}\overline{\mathbf{A}}\overline{\mathbf{B}}, \dots, \overline{\mathbf{C}}\overline{\mathbf{A}}^{L-1}\overline{\mathbf{B}}\right), \tag{9b}$$

where $\circledast$ denotes convolution operation, $\overline{\mathbf{K}} \in \mathbb{R}^L$ is the one-dimensional (1D) convolutional kernel. The combination process is detailed in Appendix B. This shows that S4 models can be transferred into the convolutional form so that they can be trained in parallel like CNNs, and the explanation of parallel training is provided in Appendix C.1. In practice, S4 models pre-compute Krylov matrix $\overline{\mathbf{K}}_{\mathfrak{k}} \in \mathbb{R}^{N\times L}$ of the convolutional kernel $\overline{\mathbf{K}}$ [17]:

$$\overline{\mathbf{K}}_{\mathfrak{k}} = \left(\overline{\mathbf{B}}, \overline{\mathbf{A}}\overline{\mathbf{B}}, \dots, \overline{\mathbf{A}}^{L-1}\overline{\mathbf{B}}\right). \tag{10}$$

Subsequently, the kernel $\overline{\mathbf{K}}$ can be calculated through a matrix multiplication: $\overline{\mathbf{K}} = \overline{\mathbf{C}}\overline{\mathbf{K}}_{\mathfrak{k}}$. Finally, Eq. $(9a)$ is executed in parallel. Besides, it is evident that S4 models can intrinsically capture long-term dependencies and bring global perception without the intricate hierarchical stacking structures found in CNNs. The aforementioned details describe the fundamental concept underlying S4 models.

Further, it becomes apparent that the primary computational burden of S4 models lies in computing the latent state $\overline{\mathbf{K}}_{\mathfrak{k}}$, which necessitates $L$ consecutive matrix multiplications by $\overline{\mathbf{A}}$, involving $O(N^2L)$ multiplication operations and $O(NL)$ space complexity [18]. The detailed complexity statements are provided in Appendix D. This substantial computational load constitutes the fundamental bottleneck of the naïve S4 models. Subsequent advancements in S4 models pay attention to how to improve computational efficiency. For example, Gu et al. [18] have discovered that $\overline{\mathbf{A}}$ can be transformed into a diagonal plus low-rank (DPLR) form, and this technique can reduce the computational complexity to $\tilde{O}(N+L)$ and the space complexity to $O(N+L)$, respectively, where $\tilde{O}(1)$ denotes the cost of a single Cauchy matrix-vector multiplication.

Gu and Dao [16] observe that conventional S4 models are all inherently linear time invariant and lack the capability to selectively focus on or disregard particular inputs because the parameters $\overline{\mathbf{K}}$ are not related to inputs once SSMs are well-trained. Therefore, to address this limitation, they introduce a selection mechanism into S4 models, giving rise to Mamba (referred to as SSMs+Selection). The core idea of the selection mechanism lies in parameterizing $\overline{\mathbf{A}}$, $\overline{\mathbf{B}}$ and $\overline{\mathbf{C}}$ in an input-dependent manner, enabling the new model to dynamically select valuable information and filter out irrelevant data indefinitely based on the inputs. Specifically, Mamba models $\mathbf{B}$, $\mathbf{C}$, and $\Delta$ as the functions of the input sequence $\mathbf{x}$:

$$\mathbf{B} = S_{\mathbf{B}}(\mathbf{x}), \mathbf{C} = S_{\mathbf{C}}(\mathbf{x}), \Delta = \tau_{\Delta}\left(Parameters + S_{\Delta}(\mathbf{x})\right), \tag{11}$$

where $S_{\mathbf{B}}(\mathbf{x}) = \text{Linear}_N(\mathbf{x})$ and $S_{\mathbf{C}}(\mathbf{x}) = \text{Linear}_N(\mathbf{x})$, with $\text{Linear}_N$ representing a linear and parameterized projection to dimension $N$. In the neural network domain, this is also named a fully connected layer (FCL). The symbol $\tau_{\Delta}$ represents the Softplus activation functions, and $S_{\Delta} = \text{Broadcast}_D\left(\text{Linear}_1(\mathbf{x})\right)$, which is inspired by the recurrent neural network (RNN) gating mechanism [16]. Regarding the matrix $\mathbf{A}$ in Eq. $(2a)$, it remains input-invariant and is initialized using the HiPPO Matrix for simplicity. This is due to $\overline{\mathbf{A}} = f_{\mathbf{A}}(\Delta, \mathbf{A})$ and $\overline{\mathbf{B}} = f_{\mathbf{B}}(\Delta, \mathbf{A}, \mathbf{B})$ (refer to Eq. $(4)$), where selectivity in $\Delta$ and $\mathbf{B}$ guarantees selectivity in $\overline{\mathbf{A}}$ and $\overline{\mathbf{B}}$. Subsequently, in the Mamba paper [16], Gu and Dao claim that they employ the zero-order hold (ZOH) technique to discretize the state space equations provided in Eqs. $(2a)$ and $(2b)$. The new parametric representations of $\overline{\mathbf{A}}$, $\overline{\mathbf{B}}$, and $\overline{\mathbf{C}}$ are

$$\overline{\mathbf{A}} = \exp(\Delta\mathbf{A}), \tag{12}$$

$$\overline{\mathbf{B}} = (\Delta\mathbf{A})^{-1}(\exp(\Delta\mathbf{A}) - \mathbf{I}) \cdot \Delta\mathbf{B}, \tag{13}$$

and $\overline{\mathbf{C}} = \mathbf{C}$. The proof of Eqs. (12) and (13) is included in Appendix A.3. However, in their official code implementation, they choose a simplified version of the ZOH discretization method to facilitate computation without affecting empirical performance*:

$$\overline{\mathbf{A}} = \exp(\Delta\mathbf{A}), \tag{14}$$

$$\overline{\mathbf{B}} = \Delta\mathbf{B}, \tag{15}$$

and $\overline{\mathbf{C}} = \mathbf{C}$. The analysis of the approximation version is given in Appendix A.4. It is critical to stress that the simplified discretization representation is adopted by almost all the Mamba-based research. † The preceding discussions constitute the first contribution of the Mamba paper [16], *i.e.*, the Mamba model.

The second contribution lies in the design of a straightforward and homogenous module incorporating the Mamba model, drawing inspiration from the gated multi-layer perceptron (MLP). The module is referred to as the vanilla Mamba block in this study and is depicted in Fig. 1 (a), where Gu and Dao choose SiLU or Swish as the nonlinear activation function. The core component of the vanilla Mamba block is the SSM module which executes the aforementioned Mamba model, which is also known as the SSMs+Selection mechanism, and henceforth denoted as SSM of Fig. 1 (a) for brevity. Recognizing the complexity of Mamba's computations, we illustrate the internal operational procedures of the SSM module in Fig. 1 (b) to facilitate readers' comprehension. The default scanning direction is the forward path, progressing from $x_0$ to $x_{L-1}$, as depicted by the direction of the Mamba snake in Fig. 1 (b).

* https://github.com/state-spaces/mamba/issues/19, or https://github.com/state-spaces/mamba/issues/114, or https://github.com/state-spaces/mamba/issues/129

† https://github.com/alxndrTL/mamba.py/issues/10

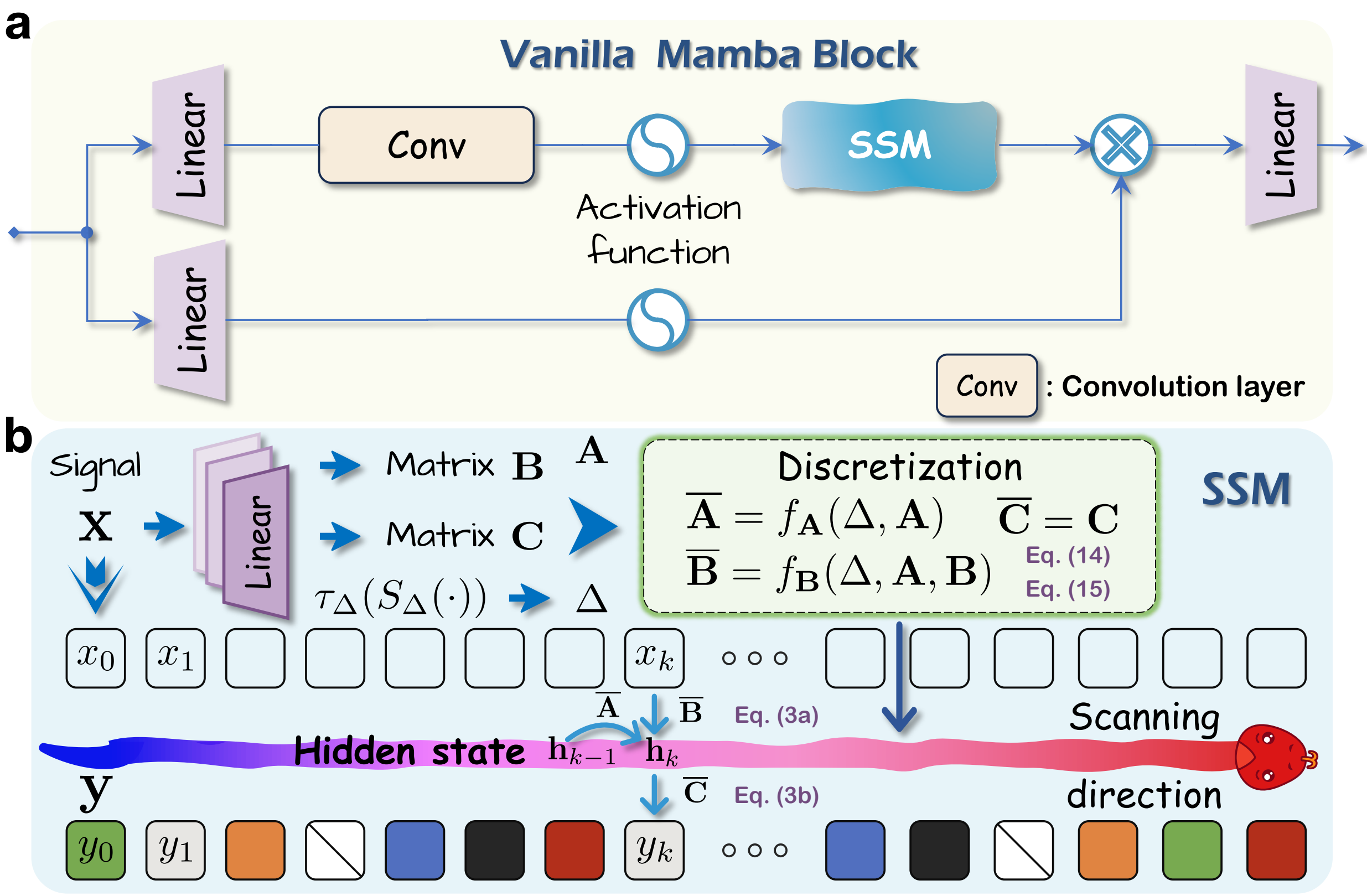

**Fig. 1.** The illustration of the vanilla Mamba block and SSM: (a) the vanilla Mamba block and (b) the internal operational procedures of SSM.

The ultimate contribution of the Mamba paper lies in the development a hardware-aware algorithm aimed at accelerating calculations. A fundamental enhancement offered by Mamba, in contrast to S4 models, is the transition of the parameters from input-invariant to input-dependent. However, every coin has two sides. The enhancement introduces a technical challenge for Mamba's computation: the state space equations in Eqs. $(3a)$ and $(3b)$ can no longer be transformed into the convolutional form as delineated in Eq. $(9a)$ (refer to Appendix C.2 for an exhaustive analysis), thereby hindering parallel computation of the equations. This is why we describe the equations in their original recurrent form in Fig. 1 (b). Considering that executing the state space equations in the recurrent representation is so time-consuming and computationally demanding, the Mamba paper employs a parallel associative scan algorithm, enabling parallel computation of Eq. $(3a)$ with a theoretical computational complexity of $O(LN)$ (scaling linearly with the input length $L$). Moreover, to enhance the speed and memory efficiency of Mamba's calculation, the Mamba paper implements the kernel fusion and the recomputation techniques, which can allow the discretization and the entire scanning process of Eqs. $(3a)$ and $(3b)$ both in SRAM (highly computational-efficient) of GPUs, thereby significantly reducing the need for data transfers between SRAM and HBM, which is adopted by the former SSMs and comparatively less efficient.

In essence, Mamba exhibits linear computational and spatial complexity scaling with sequence length and achieves $5\times$ higher throughput than Transformers, as demonstrated in practice [16]. Moreover, Mamba is inherently capable of global modeling and has demonstrated state-of-the-art (SOTA) performance across various sequence modeling tasks, such as language processing, audio analysis, and genomics modeling. Consequently, Mamba has garnered significant interest and is considered to potentially replace Transformers as the next generation backbone network.

Nevertheless, the prowess of Transformer lies in its versatility across various modalities, encompassing not only sequential information like text and audio but also visual elements such as images and videos. Therefore, researchers have promptly integrated Mamba into the realm of computer vision (CV), which is detailed in the subsequent subsection.

### *2.2. Mamba in computer vision*

Vim and VMamba take the lead to explore the effectiveness of Mamba in the field of CV, standing out as the two most prominent visual Mamba models. Thus, this study solely investigates these models for the domain of Mamba in computer vision.

Given that the scanning mechanism in the vanilla Mamba block operates along a 1D sequence (from $x_0$ to $x_{L-1}$), which contradicts the two-dimensional (2D) image space, Vim and VMamba have adopted the same solution as that of Vision Transformer (ViT) [27]. Specifically, first, represent a 2D image as $\mathbf{M} \in \mathbb{R}^{C\times H\times W}$, where $C$, $H$, and $W$ denote the channel, height, and width, respectively. Subsequently, the image is partitioned into numerous small fixed-size patches along the height and width dimensions, with each patch having a spatial resolution of $((h_P, w_P))$. This results in the shape of each patch being $\mathbb{R}^{C\times h_P\times w_P}$, and the total number of patches being $N = (HW)/(h_P w_P)$. Next, each patch is flattened into a 1D sequence, transforming its shape to $\mathbb{R}^{(C\times h_P\times w_P)}$. Consequently, the original image is reshaped into $\mathbf{M}_P \in \mathbb{R}^{N\times(C\times h_P\times w_P)}$, where $\mathbf{M}_P$ can be viewed as a 1D sequence comprising $N$ elements and each element with a dimension of $(C \times h_P \times w_P)$. Specifically, $\mathbf{M}_P[i]$, $i \in \{0,1,\dots,N-1\}$ corresponds to $x_k$ in Eq. $(3a)$. With the acquisition of the input 1D sequence, theoretically, we can direct use Mamba to model global dependencies. However, to enhance the handling of spatial context information inherent in visual data, Vim and VMamba have refined the vanilla Mamba block, with detailed enhancements introduced below.

Vim imitates the vanilla Mamba block to design the vision Mamba (Vim) block, as depicted in Fig. 2 (a). In contrast to the vanilla Mamba block (refer to Fig. 1 (a)), Vim designs a bidirectional scanning strategy illustrated in Fig. 2 (b), encompassing both a forward and a backward route. Vim executes a singular SSM operation (see Fig. 1 (b)) for each route. Leveraging this bidirectional scanning strategy, each element $\mathbf{M}_P[i]$ in the sequence amalgamates information not only from $\mathbf{M}_P[0]$ to $\mathbf{M}_P[i]$ (forward information) but also from $\mathbf{M}_P[N-1]$ to $\mathbf{M}_P[i]$ (backward information). Consequently, Vim can model long-range dependency relationships more effectively than the vanilla Mamba block. Moreover, the symbol $L$ in Fig. 2 (a) denotes the embedded count of Vim blocks, with $L$ set to 2 in the experiments conducted in this paper.

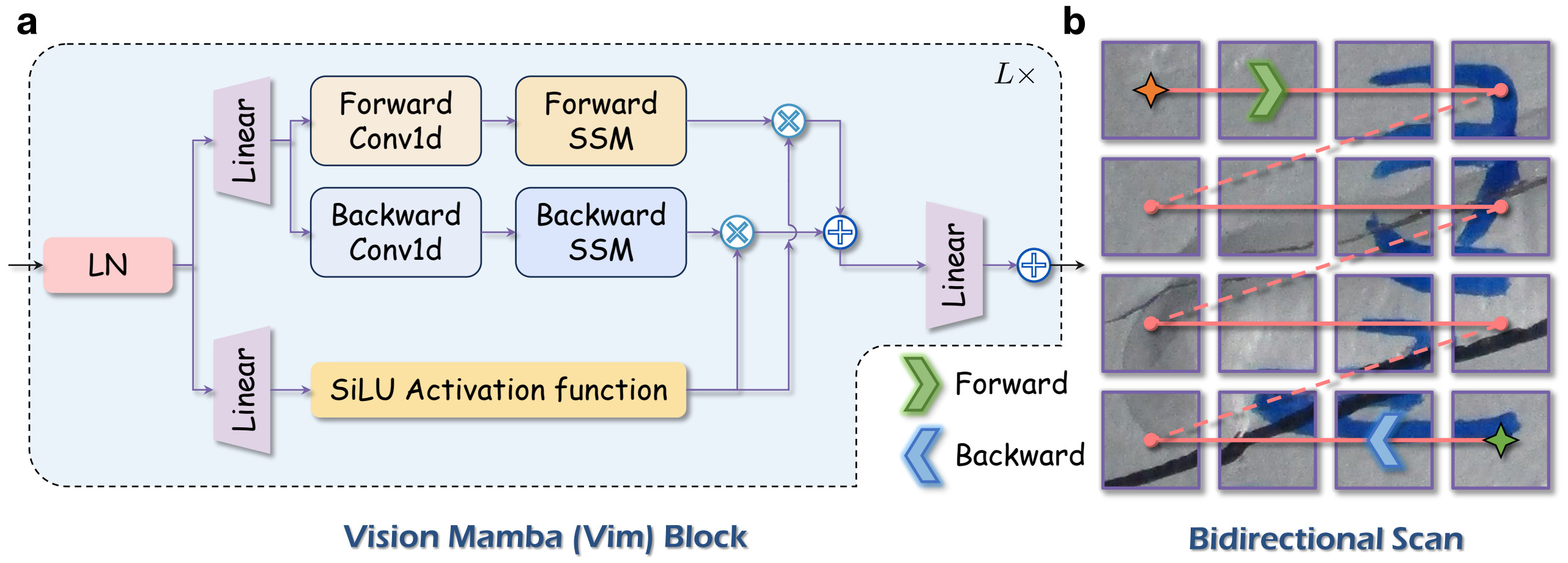

**Fig. 2.** The illustration of Vim: (a) the vision Mamba (Vim) block and (b) the bidirectional scan in the Vim block.

VMamba introduces an innovative extension by broadening the bidirectional scan into a cross scan, encompassing four distinct scanning paths, as illustrated in Fig. 3 (a). Notably, the top two routes in Fig. 3 (a) correspond to the forward and backward routes in Vim. Subsequently, VMamba designs a 2D-Selective-Scan (SS2D) module to actualize the cross scan. The SS2D module consists of two primary steps: initially, SS2D conducts a single SSM operation (as depicted in Fig. 1 (b)) for each scanning route, followed by summing the outcomes from the four scanning paths to yield its output. Moreover, VMamba designs two blocks that incorporate SS2D as the fundamental components that make up visual Mamba networks: the vanilla Visual State Space (VSS) block and the VSS block. The conceptualization of the vanilla VSS block draws inspiration from the vanilla Mamba block (refer to Fig. 1 (a)), with the key distinction being the substitution of the SSM module with SS2D, as shown in Fig. 3 (b). Furthermore, the tremendous and widespread successes of ViT and its variants in computer vision demonstrate the effectiveness of the structure of the Transformer encoder in ViT. Thus, the VSS block mimics the Transformer encoder. Specifically, the VSS block replaces the multi-head self-attention (MHSA) in the Transformer encoder with the SS2D block, as illustrated in Fig. 3 (c).

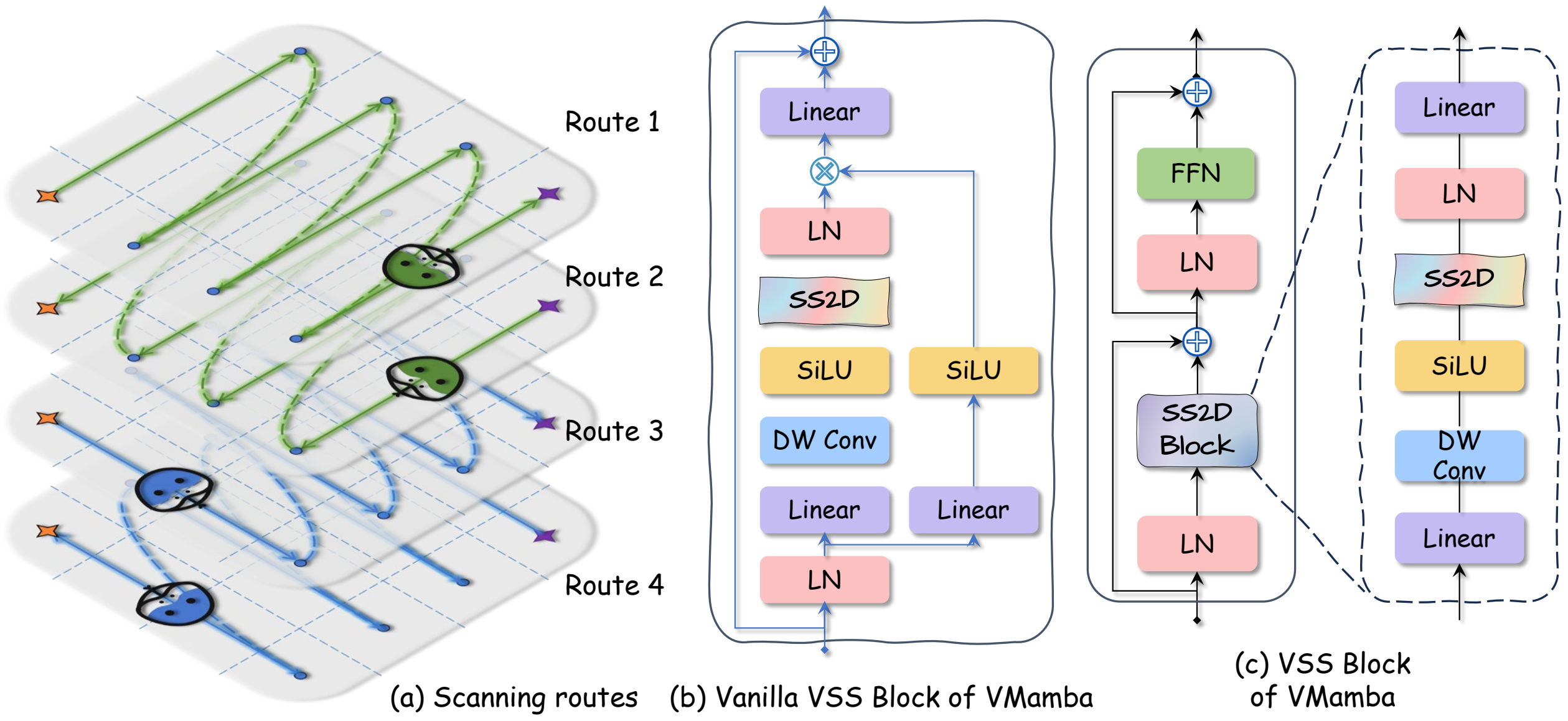

**Fig. 3.** The key components of VMamba: (a) four scanning routes of VMamba, (b) the vanilla VSS block of VMamba, and (c) the VSS block of VMamba. Note that LN denotes the layer normalization [1][28], DW Conv represents a depth-wise convolution [1], FFN is a feedforward neural network.

In the experimental section (Section 5), we select the three Mamba blocks, namely, the Vim block, the vanilla VSS block, and the VSS block to explore their representation capabilities to crack features. Theoretically, both the vanilla VSS block and the VSS block are anticipated to outperform the Vim block. This expectation arises from the SS2D module within these blocks, which incorporates four diverse scanning routes, bringing a more comprehensive information interaction and a richer spatial context information compared to the Vim block's forward and backward scanning paths. The experimental evidence presented in Table 2 and Table 4 robustly substantiates this theoretical conjecture.

Nevertheless, it is crucial to highlight that the experimental outcomes further reveal the three visual Mamba blocks cannot bring consistent positive impact to the segmentation model across distinct crack datasets. For instance, on the Deepcrack dataset, the Vim block demonstrates a decrease in performance (see Table 2), while on the Steelcrack dataset, both the Vim block and the VSS block exhibit a detrimental effect on model performance, and the vanilla VSS block showcases almost negligible impact (refer to Table 4). This observation contradicts the conclusions drawn in prior studies such as Vim [19], VMamba [20], Mamba-UNet [21] (utilizing the vanilla VSS block), or RS3Mamba [22] (also employing the vanillza VSS block), where Mamba consistently achieved remarkable performance for not only general visual representation tasks but also specific domains like medical image data and remote sensing images. Given Mamba's track record of achieving SOTA performance in various models and tasks, this study posits that the ineffectiveness of the three Mamba blocks in the crack segmentation task is not inherently linked to Mamba itself, but rather to the specific structures of these blocks. Consequently, a thorough examination of the essence of Mamba is conducted, leading to the proposal of a novel block named CrackMamba for crack segmentation, which is detailed in the subsequent subsection.

### 2.3. *Mamba block in crack segmentation*

#### 2.3.1. *Motivation: an attention perspective*

First, let us rethink Mamba from a higher and more profound perspective. The fundamental enhancement introduced by Mamba is its selection mechanism compared to the S4 models. This mechanism makes the parameters in Mamba dynamic and input-dependent, thereby eliminating the time-invariance characteristic. Essentially, this dynamic parameterization empowers Mamba-based models to intelligently select relevant information while filtering out worthless data, enhancing the model's representation power and overall performance [16].

Next, let us reflect on the concept of human attention, which plays an essential role in human perception. For human attention, individuals can actively allocate their valuable attention to subjects of interest while diminishing focus on the objects they are less interested in [23]. The attention mechanism in the deep learning domain is inspired by this innate human ability and replicates this functionality. Specifically, the attention mechanism increases or decreases the weights of the specific regions within feature tensors to accentuate or dampen the regions. As for how to ensure the regions where the weights are increased are indeed meaningful and significant, and the increased scale of the parameters (vice versa), the core strategy is to make the attention parameters input-dependent and dynamic [29], enabling the models to autonomously capture the underlying relationships between the attention parameters and the inputs.

Upon merging Mamba with the attention mechanism, a profound realization emerges: the essence of both is the input-dependent parameters. Consequently, essentially, Mamba can be viewed as a kind of attention mechanism. Notably, the attention mechanism has been extensively leveraged in the crack segmentation task, consistently enhancing the representation power to crack features and elevating identification performance of segmentation models [1][2][23][24][25]. This observation leads to the hypothesis that if the design of the Mamba block aligns with the principles governing attention blocks, it could potentially achieve consistent improvements to baseline segmentation models. Motivated by this notion, we introduce an attention block-like Mamba block, termed CrackMamba, elaborated upon in the following subsection.

#### 2.3.2. *Design of CrackMamba*

In attention blocks, a pivotal element is the attention map (AM), which contains the relative importance information among different regions within the feature maps. Taking the three most representative attention blocks as an example, namely, SENet [29] (illustrated in Fig. 4 (a)), CBAM [30] (depicted in Fig. 4 (b)), and parallel attention module (PAM) [23] (shown in Fig. 4 (c)), their AMs are derived by normalizing the feature maps to a range of $(0,1)$ through a Sigmoid function. Subsequently, another crucial feature of the attention blocks is that the outputs of these attention modules are obtained through element-wise multiplication of the original feature maps and the AMs. Therefore, the primary enhancement introduced by CrackMamba, in contrast to the conventional Mamba blocks, lies in the incorporation of AM. Besides, the outcome of CrackMamba also entails element-wise multiplication of the original feature map with the AM, as depicted in Fig. 4 (d).

The design of the specific structure of CrackMamba is twofold. First, considering that the vanilla VSS block stands out as the sole Mamba block that does not weaken model performance among the three 2D Mamba blocks (as discussed in Subsection 2.2), CrackMamba is constructed upon the foundation of the vanilla VSS block and also incorporates a DW Conv and an SS2D module. Second, drawing inspiration from the success of the parallel attention module (PAM) in the crack segmentation task [1] and its broader applications in the civil engineering domain such as bridge defect inspection [23], building structure identification [31], and structural health inspection [32], CrackMamba's design is influenced by the architecture of PAM. Specifically, in the domain of crack segmentation, the specific manifestation of PAM is the feature fusion module (FFM) within CrackSeU, with its core part illustrated in Fig. 4 (c). Hence, CrackMamba adopts a structure similar to that of FFM, featuring two branches and a skip connection. Besides, referring to the output composition of FFM, the ultimate output of CrackMamba entails the element-wise addition of the combination of the two branches and the skip connection, as shown in Fig. 4 (c) and Fig. 4 (d).

The quantitative assessments of crack identification in Section 5 unequivocally demonstrate that CrackMamba consistently enhances the baseline model's performance across both datasets, affirming the effectiveness of CrackMamba. Moreover, benefiting from the linear spatial and computational complexity of Mamba, CrackMamba-based CrackSeU-B has fewer parameters and reduced computational complexity compared to the original CrackSeU-B (as evidenced in Table 2 and Table 4). Therefore, as a simple yet effective and plug-and-play module, CrackMamba exhibits immense potential for widespread application in crack identification. Furthermore, the concept of integrating Mamba with the attention mechanism introduces a fresh paradigm for designing Mamba blocks, offering the potential to innovate existing Mamba blocks, thereby enhancing the overall performance of Mamba-based models across various computer vision tasks.

## 3. Crack segmentation network and crack datasets

### 3.1. Crack segmentation network

#### 3.1.1. Crack Segmentation U-shape network (CrackSeU)

In order to assess the effectiveness of Mamba to crack identification, we choose a representative crack segmentation network from a recent publication in a top-tier journa, the Crack Segmentation U-shape network (CrackSeU) [1]. This choice is guided by two primary considerations.

First, CrackSeU is built on the U-structure which is the most representative and the most successful structure in image segmentation networks due to its efficient feature fusion mechanism. In crack segmentation networks, the U-structure is a common presence [7][33][34], indicating that summarized patterns and insights from CrackSeU are universally applicable to a broad spectrum of crack segmentation networks. Second, the key innovation of CrackSeU lies in extending the two-level fusion paradigm of conventional U-shape networks like U-Net [35], U-Net++ [36], and Attention U-Net [37] to multi-level fusion. This multi-level fusion strategy enables more effective and thorough multi-scale integration of crack features and facilitates intra-network information flow, resulting in more superior crack segmentation performance compared to other U-shape networks [1]. In view of this, it is meaningful to investigate whether Mamba can further elevate the performance of this potent model. If Mamba proves successful in enhancing CrackSeU, it signifies the potential to deliver similar or even greater advancements in the performance of other ordinary U-shape segmentation networks.

CrackSeU consists of five encoder blocks (Stage 1~Stage 5), four upsampling blocks ($\mathrm{UB}_k, k = 2,3,4,5$), three feature fusion modules ($\mathrm{FFM}_k, k = 1,2,3$), one side output block (SOB) and one output block (OB). Based on the difference in FFM's volume, CrackSeU has three configurations: CrackSeU-B, CrackSeU-M, and CrackSeU-L, denoting base, middle, and large volume models, respectively. Considering that CrackSeU-B achieves the best balance between performance and computational complexity, CrackSeU-B is chosen as the final research object. For simplicity, in this paper, CrackSeU refers to CrackSeU-B unless specified otherwise.

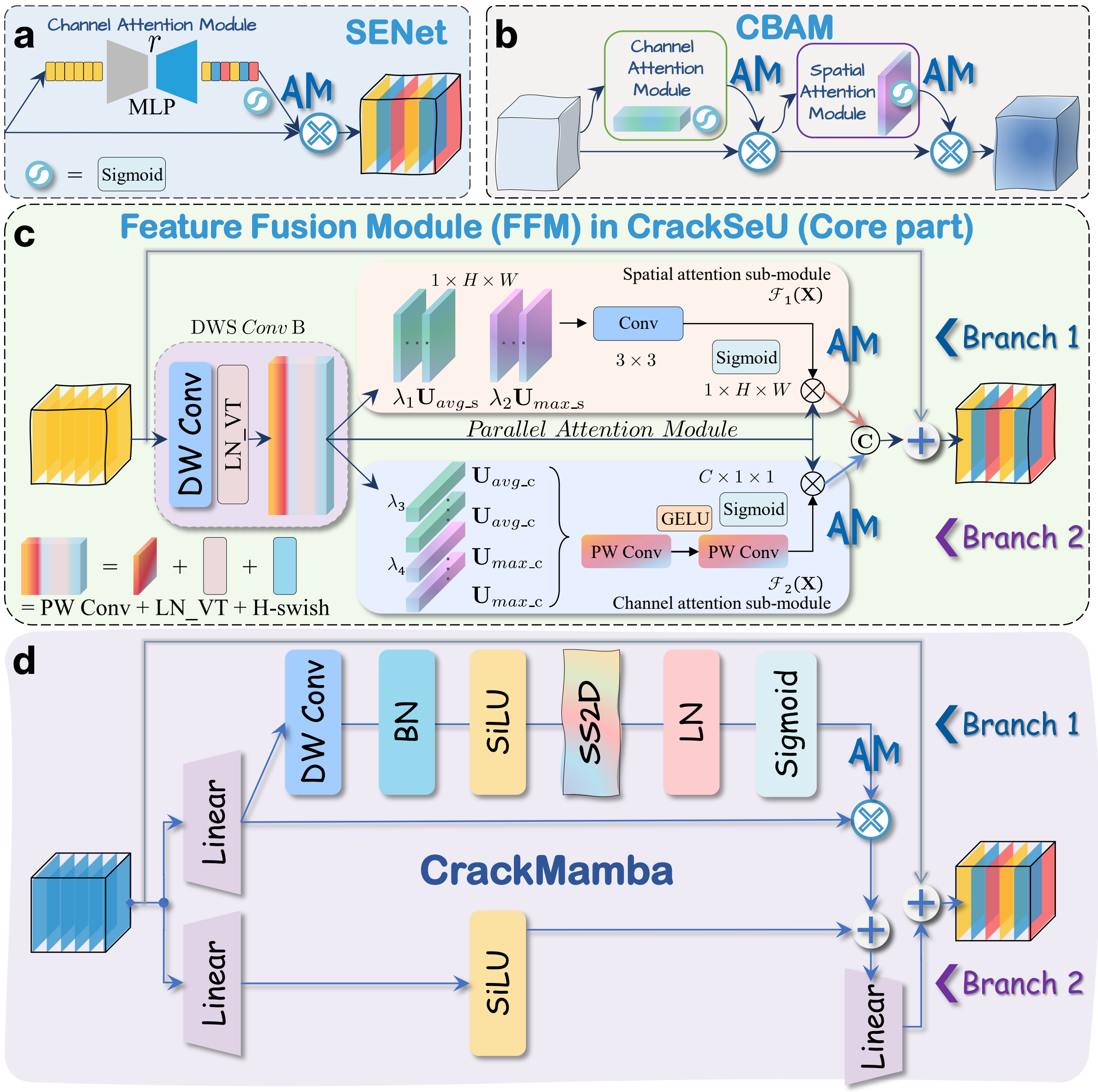

**Fig. 4.** Attention mechanism and Mamba: (a) the feature fusion module in CrackSeU and (b) the CrackMamba block. Note that PW Conv denotes a point-wise convolution, BN is the batch normalization, LN_VT is one kind of layer normalization, GELU and Sigmoid is the nonliner activation function.

### *3.1.2. Integrate CrackSeU-B and Mamba*

Next, a key challenge is how to integrate CrackSeU-B and Mamba. In this paper, we do not additionally design new fancy architectures for integration, instead, we simply substitute the original convolutional blocks in Stage 2~Stage 5 of CrackSeU-B with Mamba blocks, and the schematic diagram is shown in Fig. 5. The rationale behind this arrangement is that we want to investigate the potential and generalization of Mamba in crack identification. Clearly, if researchers conduct optimization on the structure of the integration module, the performance of the new model can be improved, however, it becomes challenging to discern whether the improvements come from Mamba or the well-crafted integration module. Moreover, a complicated module may weaken the integrated model's generalization ability, as the benefits of an elaborate structure could diminish with alterations in the training pipeline or the baseline model. Besides, it is important to highlight that the numbers of channels vary among the input features of each stage within the original CrackSeU-B. Nevertheless, the Mamba blocks are all plug-and-play modules, that preserve the shapes of input feature. To solve the contradiction, a point-wise convolution (PW Conv) is incorporated to align the output features of the Mamba blocks with the channel configurations of CrackSeU-B.

#### *3.1.3. Loss function*

To achieve fair comparisons, the loss functions adopted in this paper are consistent with the original loss functions training CrackSeU. The total loss function $\mathcal{L}_{total}$ is a mixed loss function:

$$\mathcal{L}_{total} = \alpha_1 \times \frac{1}{B}\sum_{i=1}^{B} \mathcal{L}_{\mathrm{BCE}}(\mathbf{P}^i, \mathbf{G}^i) + \alpha_2 \times \frac{1}{B}\sum_{i=1}^{B} \mathcal{L}_{\mathrm{Dice}}(\mathbf{P}^i, \mathbf{G}^i) + \alpha_3 \times \frac{1}{B}\sum_{i=1}^{B} \mathcal{L}_{\mathrm{BCE}}^{S}(\mathbf{P}_s^i, \mathbf{G}_s^i), \tag{14}$$

where $B$ is the batch size. $\alpha_1$, $\alpha_2$ and $\alpha_3$ are three hyperparameters for regulating the impact of different sub-loss functions, and they are set to 1, 1 and 0.1, respectively. $\mathbf{P}^i \in (0,1)^{1\times H\times W}$ and $\mathbf{G}^i \in \{0,1\}^{1\times H\times W}$ denote the predicted probability of the $i$-th image in a mini-batch and the corresponding ground-truth label. $\mathbf{P}_s^i \in (0,1)^{1\times H/2\times W/2}$ and $\mathbf{G}_s^i \in [0,1]^{1\times H/2\times W/2}$ represent the $i$-th side output image and the corresponding human label. It is important to note that the image size of the side output is half of the final output, as shown in Fig. 5. $\mathcal{L}_{\mathrm{BCE}}$ and $\mathcal{L}_{\mathrm{BCE}}^{S}$ are two binary cross-entropy (BCE) loss functions and they can be formulated as follows:

$$\mathcal{L}_{\mathrm{BCE}} = -\frac{1}{H\times W}\sum_{(h,w)}[\mathbf{G}(h,w)\times\log(\mathbf{P}(h,w)) + (1-\mathbf{G}(h,w))\times\log(1-\mathbf{P}(h,w))], \tag{15}$$

and

$$\mathcal{L}_{\mathrm{BCE}}^{S} = -\frac{1}{\frac{H}{2}\times\frac{W}{2}}\sum_{(h',w')}[\mathbf{G}_{\mathrm{s}}(h',w')\times\log(\mathbf{P}_{\mathrm{s}}(h',w')) + (1-\mathbf{G}_{\mathrm{s}}(h',w'))\times\log(1-\mathbf{P}_{\mathrm{s}}(h',w'))], \tag{16}$$

where $h \in [1,H]$, $w \in [1,W]$, $h' \in [1,H/2]$, and $w' \in [1,W/2]$. $\mathcal{L}_{\mathrm{Dice}}$ is a Dice loss function, which is effective to alleviate class-imbalance problems. Considering that cracks are always subtle, thus, crack/non-crack pixels are of a highly imbalanced distribution, the Dice loss function is introduced and can be defined as

$$\mathcal{L}_{\mathrm{Dice}} = 1 - \frac{2\sum_{(h,w)}[\mathbf{P}(h,w)\times\mathbf{G}(h,w)] + \epsilon}{\sum_{(h,w)}[\mathbf{P}(h,w)] + \sum_{(h,w)}[\mathbf{G}(h,w)] + \epsilon}, \tag{17}$$

where $\epsilon$ is a Laplace smoothing item to avoid zero division and is set to $1\times10^{-4}$.

### *3.2. Crack datasets*

The academic community has open-sourced numerous influential crack datasets, and these datasets usually can be categorized into two groups based on the crack type: asphalt or concrete pavement crack datasets and steel structure crack datasets. The former category accounts for the vast majority of the proportion due to its accessibility and includes CrackTree206 [38], CFD [39], CRACK500 [40], and Deepcrack [41]. In contrast, the latter category currently comprises only one publicly available dataset, namely Steelcrack [2].

To comprehensively evaluate the impact of Mamba on crack identification accuracy, we select a representative dataset from each category. The first dataset chosen is Deepcrack, which comprises 537 crack images, along with the corresponding pixel-level labels. The crack images are captured in multiple pavement scenes, which can be divided into two categories: asphalt pavements and concrete pavements. The training set of Deepcrack consists of 300 pairs of crack images and labels, while the remaining 237 pairs of images and labels form the test set. The pixel resolutions of the images are 544 × 384 or 384 × 544. Since the number of images in the original training set of Deepcrack is so limited, we conduct the same data augmentation strategies for the training set as those in [1]. Initially, all images are horizontally and vertically flipped, as well as rotated by 90°, 180°, and 270°. Subsequently, the image resolutions are resized to 480 × 480 to account for changes in aspect ratio due to rotation transformation. A random cropping strategy is then implemented, with a fixed crop size of 384 × 384. The details of the augmented Deepcrack dataset are outlined in Table 1. The second dataset selected is Steelcrack, where all the crack images are directly captured from various engineering projects of steel structures, such as Humen Bridge and Nanjing Second Yangtze River Bridge. Steelcrack also provides the corresponding meticulous manual labels for all images. All the images and labels in Steelcrack maintain a consistent pixel resolution of 512 × 512. Steelcrack consists of 3300 training images, 525 images for validation, and 530 images in the test set. As the original Steelcrack dataset has already undergone data augmentation, we do not conduct further data augmentation anymore. The specifics of the Steelcrack dataset are presented in Table 1 below.

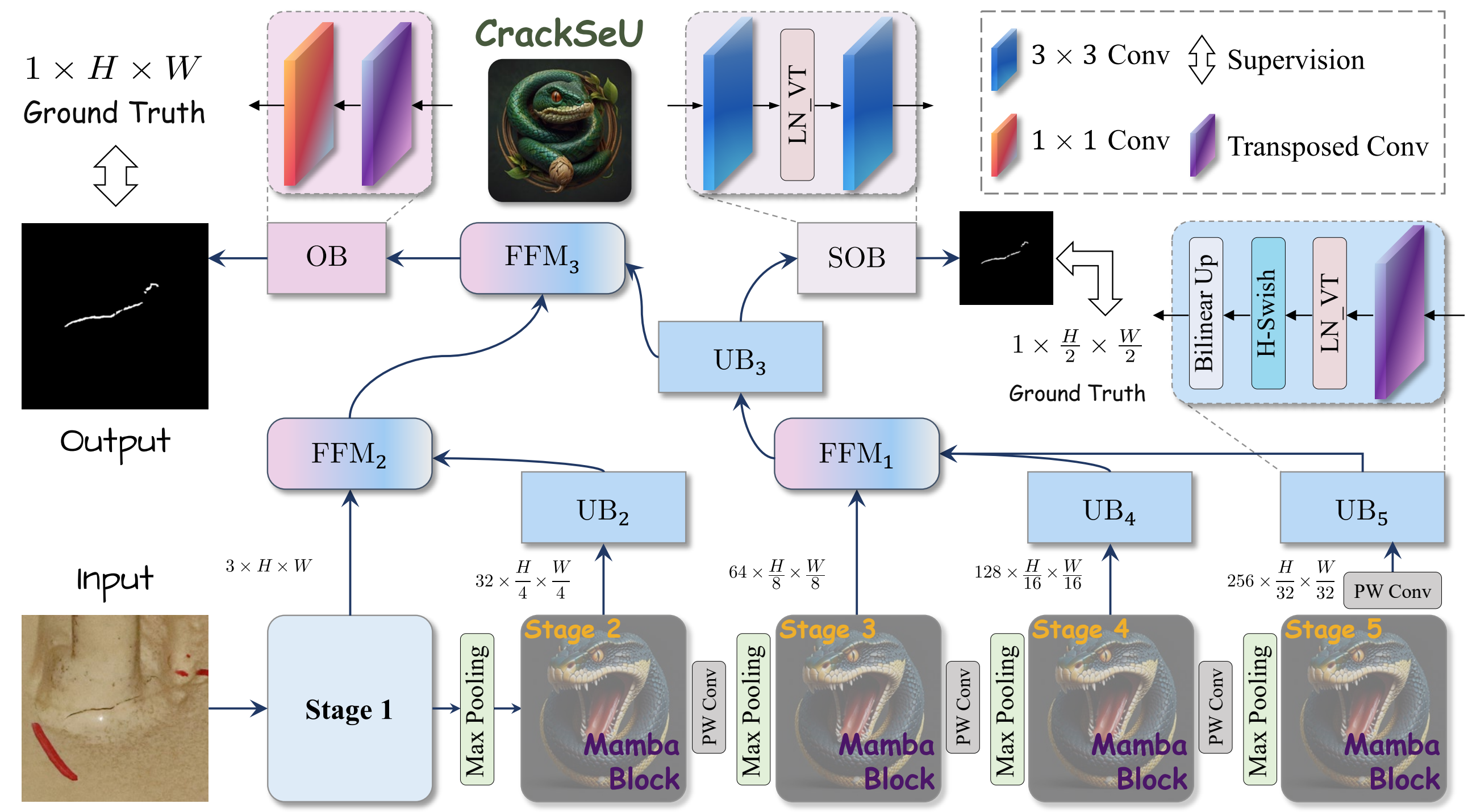

**Fig. 5.** CrackSeU-B [1] network integrating Mamba.

**Table 1** Details of the two adopted datasets.

| Dataset name | Image size for training | Number of images for training | Image size for validation and test | Number of images for validation | Number of images for test |
|---|---|---|---|---|---|
| Deepcrack [1][41] | Resized to 480 × 480 | 1800 | 544 × 384 or | 0 | 237 |
| | Crop size 384 × 384 | 1800 | 384 × 544 | | |
| Steelcrack [2] | 512 × 512 | 3300 | 512 × 512 | 525 | 530 |

## 4. Details of implementation

### *4.1. Training and inference environment*

The hardware used for training and inference includes an Intel Xeon E5-2697 @ 2.70 GHz CPU, 3 × Nvidia GeForce RTX 2080Ti GPUs, and RAM is 256 GB. The main software environment: the Ubuntu 18.04 operating system, CUDA 11.6.2, CUDNN 8.6.0, and Python 3.8.15. The deep learning framework employs version 1.12.1 of PyTorch developed by Meta.

The package versions associated with Mamba are as follows: the version of causal-conv1d is 1.1.1. The Mamba utilized in this study is a hybrid of versions 1.2.0 and 1.1.1 sourced from Vim‡. Due to installation bugs frequently encountered, we find this relatively complex installation pathway after multiple attempts. In our openly accessible code repository (https://github.com/hzlbbfrog/CrackMamba), we have provided a detailed tutorial about the installation process to bolster credibility and reproducibility.

### 4.2. Training policy

For the training policies of the two datasets, we follow the settings in the paper [1] and [2], respectively. Specifically, the training policy of Deepcrack follows [1], where the optimizer is the Adam optimizer, and the hyper-parameters are kept to the default values provided by PyTorch, *i.e.*, $\beta = (0.9, 0.999)$, $\varepsilon = 10^{-8}$ and $weight\ decay = 0$. The learning rate is fixed at $9 \times 10^{-4}$. The networks are all trained for 80 epochs before inference, and the batch size is set to 12. The Steelcrack's training policy follows the configuration in [2], where the optimizer is still the Adam optimizer, and the hyper-parameters are not changed. The learning rate is $6 \times 10^{-3}$, and the batch size is set to 9. We determine the best model by training the networks for 70 epochs.

It is worth noting that for fair and standardized comparisons and replicability, for the same dataset, the training pipeline across all the models are maintained consistently.

### 4.3. Evaluation metrics

To comprehensively and thoroughly validate and assess the performance of the models, two types of metrics are employed following [2], namely, effect and efficiency metrics, as summarized below.

#### 4.3.1. Effect metrics

The effect metrics encompass two key indicators: the mean image-wise intersection over union (mi IoU) and the mean image-wise Dice coefficient (mi Dice). Mi IoU is a discrete measure that emphasizes pixel-level similarity, while mi Dice serves as a continuous evaluation metric and is inclined to explore image-level performance [9]. Both metrics exhibit a positive correlation with model performance, with higher values indicating a closer resemblance between the prediction and the ground truth label.

Specifically, mi IoU is defined by the following formula:

$$\text{mi IoU} = \frac{1}{N}\sum_{n=1}^{N} \frac{TP^n + \varepsilon}{FN^n + FP^n + TP^n + \varepsilon} \times 100\%, \tag{18}$$

where $N$ denotes the number of images to be evaluated. For the $n$-th image, $TP^n$ represents the number of true positive pixels in the image, $FN^n$ denotes the number of false negative pixels, and $FP^n$ is the number of false positive pixels. $\varepsilon$ is a constant with a small value to prevent the denominator from being zero and is set to $10^{-6}$. The other effect measure, mi Dice, can be obtained by the following equation:

$$\text{mi Dice} = \frac{1}{N}\sum_{n=1}^{N} \frac{2|\mathbf{P}^n \otimes \mathbf{G}^n| + \varepsilon}{|\mathbf{P}^n| + |\mathbf{G}^n| + \varepsilon} \times 100\%, \tag{19}$$

where, $\mathbf{P}^n \in (0,1)^{1\times H\times W}$ denotes the $n$-th prediction, and $\mathbf{G}^n \in \{0,1\}^{1\times H\times W}$ is the corresponding $n$-th ground-truth.

#### 4.3.2. Efficiency metrics

Two efficiency metrics are employed to assess the computational complexity of models. The first indicator counts the number of parameters in a model and is denoted as #Param. The second index is multiply-accumulate operations (MACs), which is utilized to evaluate the computational expenses of models. To ensure consistency with prior studies in [1][2], we use the open-sourced tool Ptflops[§] to compute the two metrics. It is important to note that both #Param. and MACs are inversely correlated with model efficiency; hence, lower #Param. or MACs indicate higher computational efficiency.

‡ https://github.com/hustvl/Vim/tree/main/mamba-1p1p1

## 5. Experiments

Extensive experiments have been conducted to verify the effectiveness of the proposed method. Specifically, detailed quantitative comparisons are presented in Subsections 5.1, and Subsections 5.2 provides visualization results of the effective receptive field (ERF) of the models.

### *5.1. Quantitative evaluation*

As introduced in Section 2, we have chosen four visual Mamba blocks: the Vim block, the vanilla VSS block, the VSS block, and the proposed CrackMamba block to investigate their influence on crack feature comprehension. Subsequently, we substitute the convolutional blocks in the four stages of CrackSeU-B with these Mamba blocks individually and proceed to train the new Mamba-based models on the Deepcrack and Steelcrack datasets (refer to Section 3).

The quantitative evaluation of the results obtained from the four Mamba-based CrackSeU-B models and the baseline model (namely, CrackSeU-B) on the Deepcrack dataset is presented in Table 2. From the table, several key conclusions can be drawn: (1) CrackMamba achieves the best segmentation outcomes compared to the other three Mamba blocks and the baseline model; and (2) All Mamba blocks contribute to a reduction in parameters (*i.e.*, the #Param.) and computational complexity (*i.e.*, the MACs) of the baseline model, attributed to the superior efficiency of Mamba. Additionally, Table 3 provides a comparison of results between various representative segmentation networks and CrackSeU-B + CrackMamba. The comparison reveals that CrackSeU-B + CrackMamba attains the SOTA performance across all effect measures with minimal parameter quantity and computational load. Subsequently, Table 4 showcases the quantitative evaluation results of the Mamba-based CrackSeU-B models and the baseline model on the Steelcrack dataset. By combining the results from Table 2 and Table 4, it becomes evident that CrackMamba stands out as the sole Mamba block consistently delivering remarkable positive enhancements to the baseline model. Furthermore, Table 5 displays the evaluation outcomes of prominent segmentation frameworks alongside the designed CrackSeU-B + CrackMamba model. The performance comparison demonstrates that CrackSeU-B + CrackMamba ranks second only to the latest BGCrack, with marginally lower mi IoU and mi Dice values compared to BGCrack (the mi IoU and the mi Dice of CrackSeU-B + CrackMamba are 0.90 and 0.45 less than those of BGCrack, respectively). Notably, CrackSeU-B + CrackMamba surpasses all other models across all effect metrics. It is crucial to emphasize that as presented in Table 5, in terms of efficiency metrics, CrackSeU-B + CrackMamba uses the fewest parameters (79% of BGCrack) and the lowest computational demands (63% of BGCrack). Therefore, CrackSeU-B + CrackMamba actually emerges as a synthetic optimal model.

**Table 2** Quantitative evaluation of the results obtained from different Mamba-based CrackSeU-B models on Deepcrack [1][41].

| Method | mi IoU | mi Dice | #Param. | MACs |
|---|---|---|---|---|
| Without Mamba (Original CrackSeU-B) [1] | 71.71 | 81.40 | 3.19M | 8.94G |
| + Vim block [19] | 70.79 (-0.92) | 80.58 (-0.82) | 2.36M (-0.83) | 7.60G (-1.34) |
| + vanilla VSS block [20] | 71.90 (+0.19) | 81.49 (+0.09) | 1.83M (-1.36) | 7.94G (-1.00) |
| + VSS block [20] | 72.46 (+0.75) | 82.26 (+0.86) | 2.35M (-0.84) | 8.26G (-0.68) |
| + CrackMamba | 73.12 (+1.41) | 82.50 (+1.10) | 1.83M (-1.36) | 7.94G (-1.00) |

Note:

§ https://github.com/sovrasov/flops-counter.pytorch

(1) MACs are computed assuming the input images with a resolution of 3 × 544 × 384.

**Table 3** Quantitative evaluation of the results obtained from different models on Deepcrack [1][41].

| Method | mi IoU | mi Dice | #Param. | MACs |
|---|---|---|---|---|
| U-Net [35] | 68.17 | 75.07 | 7.77M | 43.84G |
| U-Net (large) | 68.40 | 75.64 | 31.04M | 174.53G |
| U-Net++ [36] | 67.92 | 74.91 | 9.16M | 110.47G |
| Attention U-Net [37] | 69.19 | 75.11 | 34.88M | 212.40G |
| CE-Net [42] | 68.80 | 76.10 | 29.00M | 28.37G |
| DeepLabv3+ (MobileNetv2) [43] | 69.18 | 74.23 | 5.81M | 23.25G |
| DeepLabv3+ (ResNet-101) [43] | 67.52 | 73.82 | 59.34M | 70.80G |
| CrackSeU-B [1] | 71.71 | 81.40 | 3.19M | 8.94G |
| CrackSeU-B + CrackMamba | 73.12 | 82.50 | 1.83M | 7.94G |

Notes:

(1) MACs are computed assuming the input images with a resolution of 3 × 544 × 384.

(2) U-Net has the number of network channels from 32 to 512, and U-Net (large) has the number of network channels from 64 to 1024. These are two commonly used configurations.

**Table 4** Performance comparison of different Mamba-based CrackSeU-B models on Steelcrack [2].

| Method | mi IoU | mi Dice | #Param. | MACs |
|---|---|---|---|---|
| Without Mamba (Original CrackSeU-B) [1] | 70.42 | 80.50 | 3.19M | 11.22G |
| + Vim block [19] | 69.41 (-1.01) | 79.60 (-0.90) | 2.36M (-0.83) | 9.54G (-1.68) |
| + vanilla VSS block [20] | 70.46 (+0.04) | 80.51 (+0.01) | 1.83M (-1.36) | 9.96G (-1.26) |
| + VSS block [20] | 69.96 (-0.46) | 80.13 (-0.37) | 2.35M (-0.84) | 10.37G (-0.85) |
| + CrackMamba | 76.26 (+5.84) | 84.88 (+4.38) | 1.83M (-1.36) | 9.97G (-1.25) |

Note:

(1) MACs are calculated assuming the input resolution is 3 × 512 × 512.

**Table 5** Performance comparison of different methods on Steelcrack [2].

| Method | mi IoU | mi Dice | #Param. | MACs |
|---|---|---|---|---|
| U-Net [35] | 68.49 | 75.13 | 7.77M | 55.01G |
| U-Net (large) | 69.81 | 76.85 | 31.04M | 219.01G |
| U-Net++ [36] | 72.23 | 78.37 | 9.16M | 138.63G |
| Attention U-Net [37] | 71.25 | 77.54 | 34.88M | 266.54G |
| CE-Net [42] | 76.00 | 81.54 | 29.00M | 35.60G |
| DeepLabv3+ (MobileNetv2) [43] | 68.22 | 71.07 | 5.81M | 29.13G |
| DeepLabv3+ (Xception) [43] | 67.40 | 71.48 | 54.70M | 83.14G |
| DeepLabv3+ (ResNet-101) [43] | 69.04 | 69.45 | 59.34M | 88.84G |
| SCRN [44] | 73.23 | 78.91 | 25.23M | 31.92G |
| TransUNet [45] | 64.34 | 72.55 | 67.87M | 129.96G |
| CrackSeU-B [1] | 70.42 | 80.50 | 3.19M | 11.22G |
| CrackSeU-L [1] | 71.66 | 81.24 | 4.62M | 28.22G |
| DconnNet [46] | 74.73 | 83.40 | 28.38M | 24.79G |
| BGCrack [2] | 77.16 | 85.33 | 2.32M | 15.76G |
| CrackSeU-B + CrackMamba | 76.26 | 84.88 | 1.83M | 9.97G |

Notes:

(1) MACs are calculated assuming the input resolution is 3 × 512 × 512.

(2) U-Net has the number of network channels from 32 to 512, and U-Net (large) has the number of network channels from 64 to 1024. These are two commonly used configurations.

### 5.2. Visual explainability

As delineated in Section 2, the fundamental role of Mamba is to model global dependency relationships just like Transformer, and Section 2 elaborates on why Mamba can achieve global receptive fields through theoretical calculations. This subsection delves into the global modeling ability of Mamba from the perspective of visual interpretability. Specifically, this study employs the visualization tool proposed in RepLKNet [47] to visualize the effective receptive fields (ERFs) of feature maps output from Stage 2 to Stage 5. The visualized outcomes are presented in Fig. 6 and Fig. 7. It is crucial to emphasize that in a visualization image of ERF, the value assigned to each pixel denotes the contribution of the unit (occupying the same spatial position as the pixel in the visualization image) from the input image to the central point in the output feature map [47]. In essence, a higher value indicates a more substantial contribution of the unit to the central point. In other words, the higher the value, the stronger the perception of the central point in the output feature map to the spatial position of the input image. Further details on ERFs and their visualization images are expounded in Appendix E. Unlike Ding et al. [47], who randomly selected 50 images from ImageNet to obtain the visualization images of ERFs, we utilize all 3300 images in the training set of Steelcrack to yield more precise visualization outcomes.

The visualization outcomes of the original CrackSeU-B are depicted in Fig. 6. From the visualization results, it is evident that CrackSeU-B can enlarge the ERFs of feature maps through training. However, post-training, the ERFs remain confined and fail to encompass a global scope. In Fig. 7, the visualized ERFs of the feature maps coming from the four stages in CrackSeU-B + CrackMamba are presented. Combining Fig. 6 and Fig. 7, a notable observation emerges: while the ERFs of CrackSeU-B + CrackMamba are initially smaller than those of CrackSeU-B before training, the former undergo significant expansion during training, ultimately achieving a global perception. This empirical evidence provides a factual basis for the global modelling capability of Mamba.

Moreover, the visualization images of ERFs post-training in Fig. 7 reveals that the pixels within the cross-shaped region surrounding the central area exhibit higher values compared to other regions. This observation indicates that the central points of the output feature maps possess a stronger perception towards the cross-shaped areas. This empirical finding aligns with theoretical analysis, which is illustrated below. The pivotal module within CrackMamba is the SS2D module, which includes four scanning routes, and the output of SS2D entails the addition of results from these four routes, as depicted in Fig. 8 (a). Consequently, the central block in Fig. 8 (a) can assimilate information from all other blocks simultaneously, which bring the central block global modeling capability. Additionally, the state space equations presented in Eqs. $(3a)$ and $(3a)$ are the recurrent form, where the essence lies in compressing sequence information into the latent state $\mathbf{h}$. Fig. 8 (b) restates the state space equations, revealing that for an output $y_k$, newer inputs (such as $x_k$ and $x_{k-1}$) undergo lower compression rates, while older inputs (like $x_0$ and $x_1$) experience higher compression rates. As a result, newer inputs exert a more substantial contribution or impact on the output. In other words, in the visualization image of ERF, the values of the pixels sharing the same spatial position as the newer inputs (namely, their indexes closer to the output index) are larger. In Fig. 8 (a), we have highlighted the blocks closer to the central blocks in blue across the four routes. Combining the four routes, we can see that the closer blocks form a cross-shaped areas, therefore, the cross-shaped areas have larger values in the visualization image of ERF (refer to Fig. 7). Notably, for the output $y_k$, the input with the same index, $x_k$, holds the greatest impact to it, thereby attributing the central area with the highest value in a visualization image of ERF. This conclusion is also consistent with the observation from the visualization results depicted in Fig. 7.

In conclusion, the visualization results align seamlessly with the theoretical analysis.

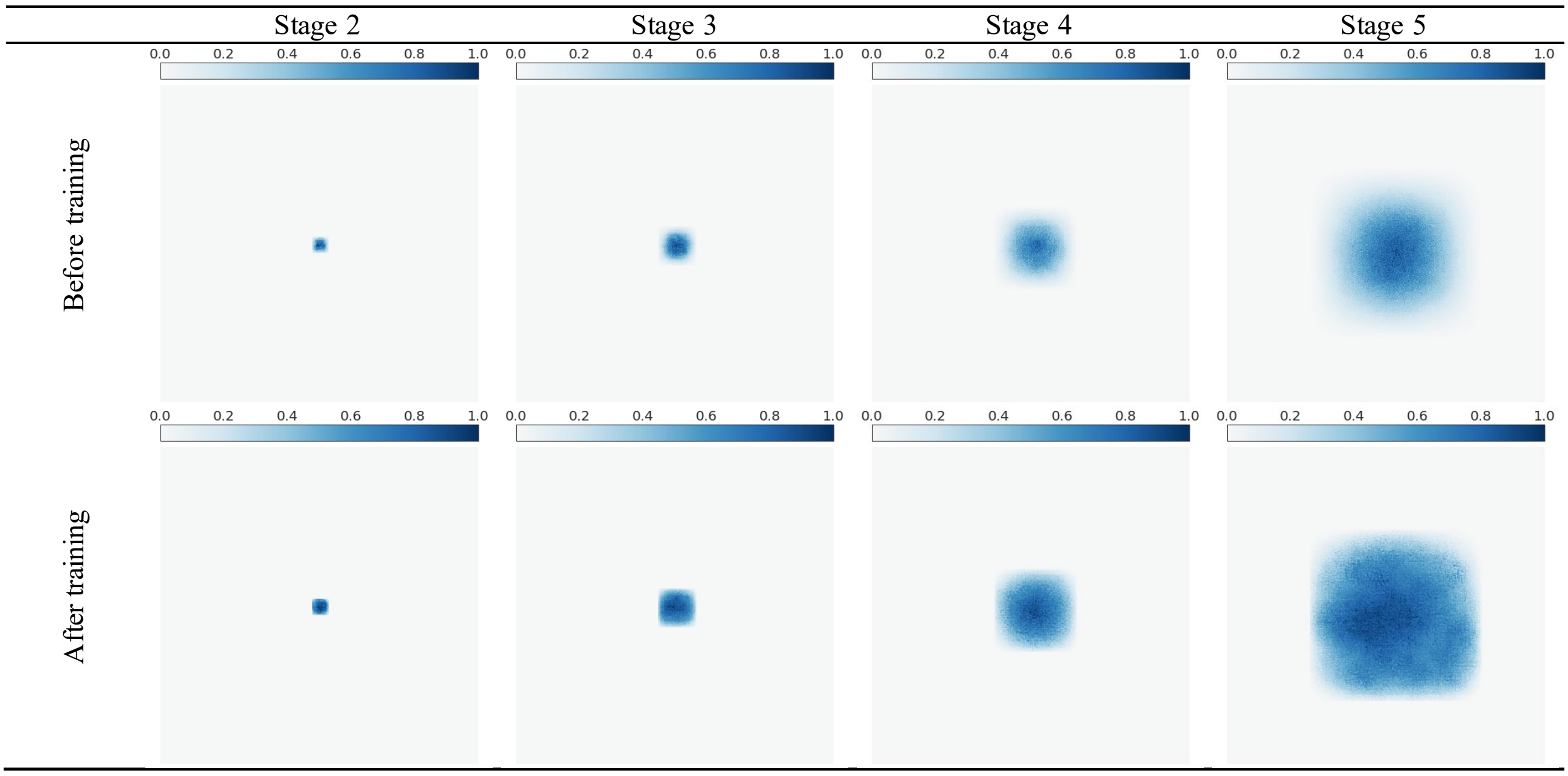

**Fig. 6.** Visualization of the effective receptive field (ERF) of CrackSeU-B.

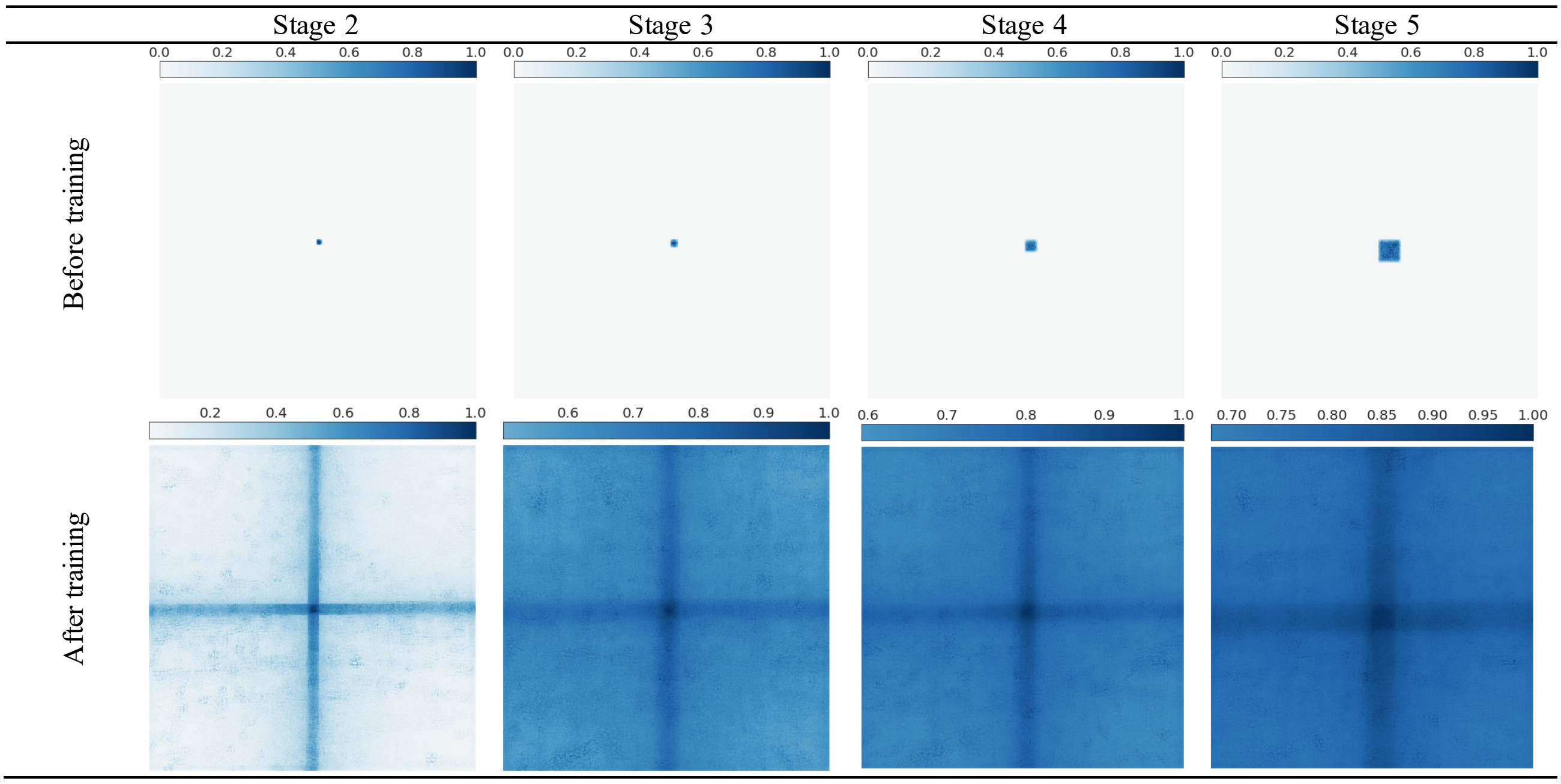

**Fig. 7.** Visualization of the effective receptive field (ERF) of CrackSeU-B + CrackMamba.

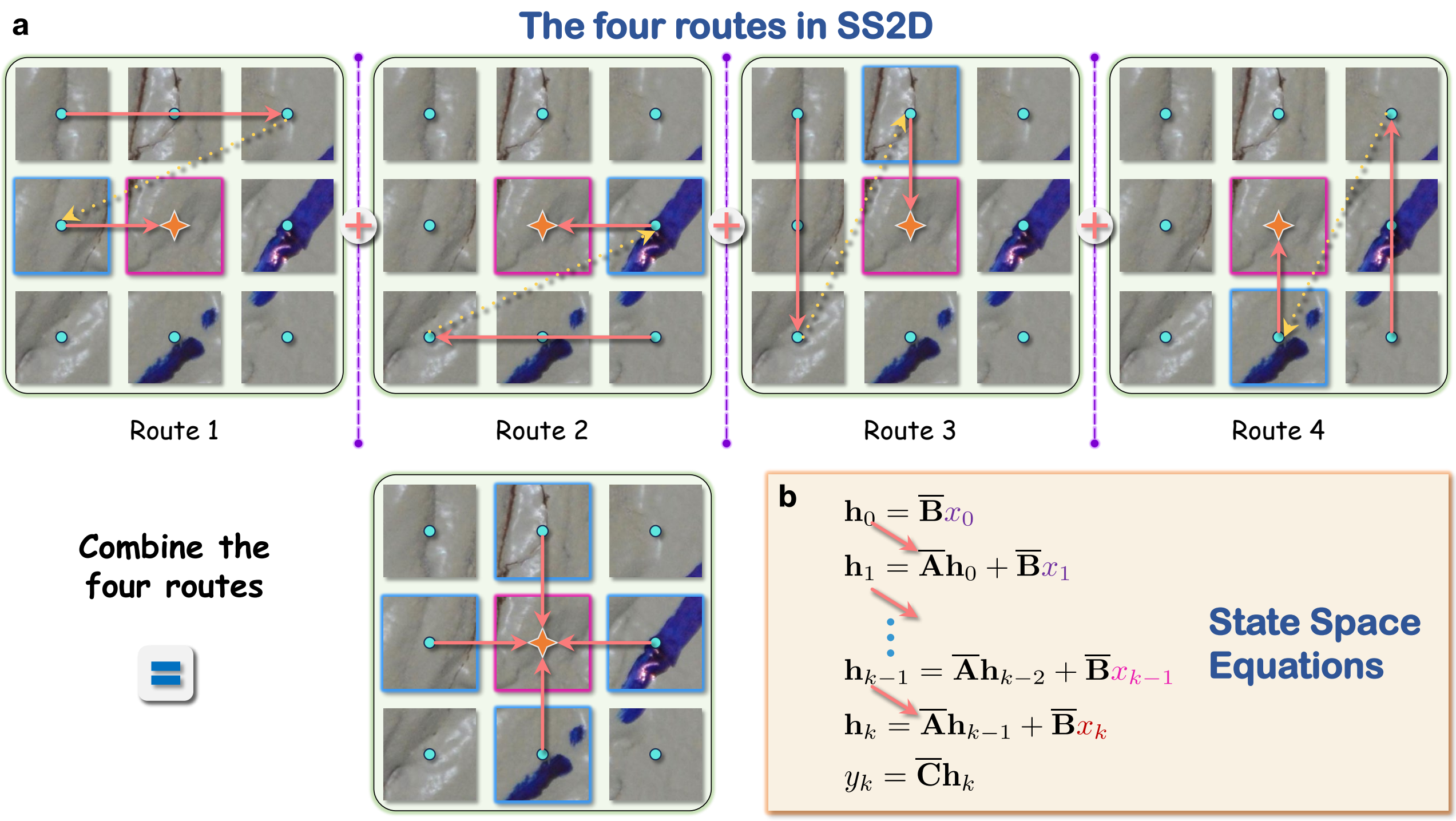

**Fig. 8.** The illustration of the visualization images of effective receptive fields (ERFs): (a) the four scanning routes in SS2D and (b) the state space equations.

## 6. Conclusions

Cracks serve as pivotal indicators of deterioration of structural performance, making the enhancement of crack segmentation precision a significant research topic. CNNs and Transformers have always been the mainstream architectures in crack segmentation networks. However, CNNs lack essential global modelling ability required for effectively capturing crack features. On the other hand, Transformers can capture long-range dependencies but come with high computational complexity and significant GPU memory requirements. Recently, the emergence of a novel architecture, Mamba, with its linear complexity and powerful global perception, has attracted considerable attention and demonstrated success across multiple CV tasks. However, its capability in learning visual representations of crack data remains unexplored. Thus, this study takes the lead to the investigation of integrating Mamba into the crack segmentation task. The principal findings and conclusions of this study are outlined below:

First, this study offers a comprehensive exposition on Mamba and vision Mamba, including Vim and VMamba as the most representative visual Mamba modules, while delving into technical details typically omitted in existing Mamba literature, aiming to facilitate understanding within the civil engineering community. Moreover, this paper introduces Vim and VMamba into the domain of crack segmentation. Specifically, these visual Mamba modules are integrated into a prominent crack segmentation model, CrackSeU, and evaluated on two diverse crack datasets, consisting of the crack images from different infrastructure domains. Surprisingly, the experimental outcomes reveal that Vim and VMamba fail to consistently enhance and, in some cases, even diminish the baseline model's performance. Subsequently, we provide a deeper insight, an attention perspective, to interpret Mamba and then design a novel Mamba module following the principles of the attention mechanism, CrackMamba. The quantitative evaluations demonstrate that CrackMamba consistently and effectively enhances the performance of the baseline model while concurrently reducing parameters and computational overhead. Serving as a plug-and-play and simple yet effective Mamba module, CrackMamba showcases enormous potential for integration into various crack segmentation models. Furthermore, the innovative Mamba design concept of combining Mamba and the attention mechanism offers significant reference value for all Mamba-based CV models, extending beyond the scope of crack segmentation networks. Last, this paper demonstrates that Mamba significantly enlarges the receptive fields of baseline models and can achieve global perception, from both the perspectives of theoretical analysis and visual explainability.

Still, further studies are recommended in the following areas. First, it is noteworthy that the Mamba-based networks in this study represent a straightforward integration of Mamba blocks with CrackSeU. Hence, researchers are encouraged to delve into further optimization of Mamba blocks and the integration methodology with crack segmentation networks to elevate performance levels. Second, this study highlights the advantageous impact of Mamba on the crack identification task. We hope this research can inspire researchers to explore more possibilities regarding the integration of Mamba within civil engineering, for example, the identification of other structural defects like spalling [23], rebar exposure, and efflorescence [48], as well as the analysis of time sequence data, such as the analysis of seismic response [49] or wind-induced vibration [50], considering that Mamba is initially proposed to sequence modeling tasks.

## Declaration of competing interest

The authors declare that they have no known competing financial interests or personal relationships that could have appeared to influence the work reported in this paper.

## Acknowledgements

The research presented was financially supported by the Innovation Technology Fund, Midstream Research Programme for Universities [project no. MRP/003/21X], and the Hong Kong Research Grants Council [project no. 16205021]. Finally, contributions by the anonymous reviewers are also highly appreciated.

## Appendix A Discretization method of ordinary differential equations

### A.1 Euler discretization method

Let's restate the continuous state space equations Eqs. $(2a)$ and $(2b)$ for convenience

$$\dot{\mathbf{h}}(t) = \mathbf{A}\mathbf{h}(t) + \mathbf{B}x(t), \tag{A1a}$$

$$y(t) = \mathbf{C}\mathbf{h}(t), \tag{A1b}$$

First, we introduce the most straightforward discretization method, the differential discretization method, also named the Euler discretization method, to convey the fundamental concept of discretizing ordinary differential equations (ODEs).

The differential discretization method. Based on the definition of derivative, we can first obtain

$$\dot{\mathbf{h}}(t) = \lim_{\Delta \to 0} \frac{\mathbf{h}(t+\Delta) - \mathbf{h}(t)}{\Delta}, \tag{A2}$$

where $\Delta \in \mathbb{R}^{+}$ denotes the step size. Next, we let $t_k = t$ and $t_{k+1} = t + \Delta$. The above equation can be transformed into

$$\dot{\mathbf{h}}(t_k) = \lim_{\Delta \to 0} \frac{\mathbf{h}(t_{k+1}) - \mathbf{h}(t_k)}{\Delta}. \tag{A3}$$

Then, we can get an approximate expression of $\dot{\mathbf{h}}(t_k)$:

$$\dot{\mathbf{h}}(t_k) \approx \frac{\mathbf{h}(t_{k+1}) - \mathbf{h}(t_k)}{\Delta}. \tag{A4}$$

Plug the above formulation into Eq. $(A1a)$:

$$\frac{\mathbf{h}(t_{k+1})-\mathbf{h}(t_k)}{\Delta}=\mathbf{A}\mathbf{h}(t_k)+\mathbf{B}x(t_k). \tag{A5}$$

Change its form:

$$\mathbf{h}(t_{k+1})-\mathbf{h}(t_k)=\Delta\mathbf{A}\mathbf{h}(t_k)+\Delta\mathbf{B}x(t_k). \tag{A6}$$

Subsequently, we can obtain the representation of $\mathbf{h}(t_{k+1})$:

$$\mathbf{h}(t_{k+1})=(\Delta\mathbf{A}+\mathbf{I})\mathbf{h}(t_k)+\Delta\mathbf{B}x(t_k), \tag{A7}$$

where $\mathbf{I}$ represents the identity matrix. We utilize $\mathbf{h}_k$ to denoted $\mathbf{h}(t_k)$ and $x_k$ to represent $x(t_k)$, then the above equation can be changed into

$$\mathbf{h}_{k+1}=(\Delta\mathbf{A}+\mathbf{I})\mathbf{h}_k+\Delta\mathbf{B}x_k. \tag{A8}$$

Therefore, recalling Eqs. $(3a)$ and $(3b)$, we have $\overline{\mathbf{A}}=\Delta\mathbf{A}+\mathbf{I}$, $\overline{\mathbf{B}}=\Delta\mathbf{B}$, and $\bar{\mathbf{C}}=\mathbf{C}$.

The Euler discretization method is straightforward enough; however, the drawback is that its error is relatively large. Therefore, S4 models utilize the bilinear discretization method.

### A.2 Bilinear discretization method

The bilinear discretization method. Unlike the differential discretization method, which is built on the derivative, this method starts with integration. First, for a function $\mathbf{h}(t)$, its integration within an interval $[t_k, t_{k+1}]$ can be approximatively formulated as

$$\int_{t_k}^{t_{k+1}}\mathbf{h}(t)\,dt\approx\frac{1}{2}[\mathbf{h}(t_k)+\mathbf{h}(t_{k+1})](t_{k+1}-t_k). \tag{A9}$$

Then, we can substitute $\mathbf{h}(t)$ with $\dot{\mathbf{h}}(t)$:

$$\int_{t_k}^{t_{k+1}}\dot{\mathbf{h}}(t)\,dt\approx\frac{1}{2}[\dot{\mathbf{h}}(t_k)+\dot{\mathbf{h}}(t_{k+1})](t_{k+1}-t_k). \tag{A10}$$

When $t_{k+1}$ close to $t_k$, we can think

$$\int_{t_k}^{t_{k+1}}\dot{\mathbf{h}}(t)\,dt=\frac{1}{2}[\dot{\mathbf{h}}(t_k)+\dot{\mathbf{h}}(t_{k+1})](t_{k+1}-t_k). \tag{A11}$$

Additionally, we know

$$\int_{t_k}^{t_{k+1}}\dot{\mathbf{h}}(t)\,dt=\mathbf{h}(t)\Big|_{t_k}^{t_{k+1}}=\mathbf{h}(t_{k+1})-\mathbf{h}(t_k). \tag{A12}$$

Combining the above two equations:

$$\frac{1}{2}[\dot{\mathbf{h}}(t_k)+\dot{\mathbf{h}}(t_{k+1})](t_{k+1}-t_k)=\mathbf{h}(t_{k+1})-\mathbf{h}(t_k). \tag{A13}$$

Based on Eq. $(\mathrm{A}1a)$, we can obtain

$$\dot{\mathbf{h}}(t_k)=\mathbf{A}\mathbf{h}(t_k)+\mathbf{B}x(t_k), \tag{A14$a$}$$

$$\dot{\mathbf{h}}(t_{k+1})=\mathbf{A}\mathbf{h}(t_{k+1})+\mathbf{B}x(t_{k+1}). \tag{A14$b$}$$

Embed Eqs. $(\mathrm{A}14a)$ and $(\mathrm{A}14b)$ into Eq. $(\mathrm{A}13)$:

$$\frac{1}{2}[\mathbf{A}\mathbf{h}(t_k)+\mathbf{B}x(t_k)+\mathbf{A}\mathbf{h}(t_{k+1})+\mathbf{B}x(t_{k+1})](t_{k+1}-t_k)=\mathbf{h}(t_{k+1})-\mathbf{h}(t_k). \tag{A15}$$

Let $\Delta = t_{k+1}-t_k$, then

$$\mathbf{h}(t_{k+1})-\mathbf{h}(t_k)=\frac{1}{2}\Delta\mathbf{A}\mathbf{h}(t_k)+\frac{1}{2}\Delta\mathbf{B}x(t_k)+\frac{1}{2}\Delta\mathbf{A}\mathbf{h}(t_{k+1})+\frac{1}{2}\Delta\mathbf{B}x(t_{k+1}). \tag{A16}$$

Namely,

$$\left(\mathbf{I}-\frac{1}{2}\Delta\mathbf{A}\right)\mathbf{h}(t_{k+1})=\left(\mathbf{I}+\frac{1}{2}\Delta\mathbf{A}\right)\mathbf{h}(t_k)+\frac{1}{2}\Delta\mathbf{B}x(t_k)+\frac{1}{2}\Delta\mathbf{B}x(t_{k+1}). \tag{A17}$$

When $\Delta$ is small enough, we can think $x(t_k)\approx x(t_{k+1})$. Thus, the above equation can be simplified as

$$\left(\mathbf{I}-\frac{1}{2}\Delta\mathbf{A}\right)\mathbf{h}(t_{k+1})=\left(\mathbf{I}+\frac{1}{2}\Delta\mathbf{A}\right)\mathbf{h}(t_k)+\Delta\mathbf{B}x(t_k). \tag{A18}$$

Change its form:

$$\mathbf{h}(t_{k+1})=\left(\mathbf{I}-\frac{1}{2}\Delta\mathbf{A}\right)^{-1}\left(\mathbf{I}+\frac{1}{2}\Delta\mathbf{A}\right)\mathbf{h}(t_k)+\left(\mathbf{I}-\frac{1}{2}\Delta\mathbf{A}\right)^{-1}\Delta\mathbf{B}x(t_k). \tag{A19}$$

We utilize $\mathbf{h}_k$ to denoted $\mathbf{h}(t_k)$ and $x_k$ to represent $x(t_k)$, then the above equation can be changed into

$$\mathbf{h}_{k+1}=\left(\mathbf{I}-\frac{1}{2}\Delta\mathbf{A}\right)^{-1}\left(\mathbf{I}+\frac{1}{2}\Delta\mathbf{A}\right)\mathbf{h}_k+\left(\mathbf{I}-\frac{1}{2}\Delta\mathbf{A}\right)^{-1}\Delta\mathbf{B}x_k. \tag{A20}$$

Therefore, recalling Eqs. $(3a)$ and $(3b)$, we have $\bar{\mathbf{A}}=(\mathbf{I}-\Delta/2\cdot\mathbf{A})^{-1}(\mathbf{I}+\Delta/2\cdot\mathbf{A})$, $\bar{\mathbf{B}}=(\mathbf{I}-\Delta/2\cdot\mathbf{A})^{-1}\Delta\mathbf{B}$, and $\bar{\mathbf{C}}=\mathbf{C}$.

### A.3 Zero-order hold (ZOH) discretization method

Before introducing the derivation, a key theorem is proved first.

**Lemma 1.** For $\mathbf{A}\in\mathbb{R}^{N\times N}$, and $t$ is a scalar, we have

$$\mathbf{A}e^{\mathbf{A}t}=e^{\mathbf{A}t}\mathbf{A}.$$

*Proof.* First, let us review the Taylor series expansion:

$$e^{at}=1+\frac{at}{1!}+\frac{(at)^2}{2!}+\frac{(at)^3}{3!}+\cdots, \tag{A21}$$

where $a$ is a constant scalar. Then, if the coefficient of $t$ is a constant matrix, the above equation is changed to

$$e^{\mathbf{A}t}=\mathbf{I}+\frac{\mathbf{A}t}{1!}+\frac{(\mathbf{A}t)^2}{2!}+\frac{(\mathbf{A}t)^3}{3!}+\cdots, \tag{A22}$$

where $\mathbf{A}\in\mathbb{R}^{N\times N}$, and $\mathbf{I}\in\{0{,}1\}^{N\times N}$ is an identity matrix.

Therefore,

$$\begin{aligned}\mathbf{A}e^{\mathbf{A}t}&=\mathbf{A}\left(\mathbf{I}+\frac{\mathbf{A}t}{1!}+\frac{(\mathbf{A}t)^2}{2!}+\frac{(\mathbf{A}t)^3}{3!}+\cdots\right)\\&=\mathbf{AI}+\frac{\mathbf{A}^2t}{1!}+\frac{\mathbf{A}^3t^2}{2!}+\frac{\mathbf{A}^4t^3}{3!}+\cdots,\end{aligned} \tag{A23}$$

and

$$\begin{aligned}e^{\mathbf{A}t}\mathbf{A}&=\left(\mathbf{I}+\frac{\mathbf{A}t}{1!}+\frac{(\mathbf{A}t)^2}{2!}+\frac{(\mathbf{A}t)^3}{3!}+\cdots\right)\mathbf{A}\\&=\mathbf{IA}+\frac{\mathbf{A}^2t}{1!}+\frac{\mathbf{A}^3t^2}{2!}+\frac{\mathbf{A}^4t^3}{3!}+\cdots,\end{aligned} \tag{A24}$$

Because $\mathbf{I}$ is an identity matrix, $\mathbf{AI}=\mathbf{IA}$. Combining Eq. (A23) and Eq. (A24), we have $\mathbf{A}e^{\mathbf{A}t}=e^{\mathbf{A}t}\mathbf{A}$. □

**Theorem 1.** The solution of the ODE formulated by Eq. $(\mathrm{A}1a)$ is

$$\mathbf{h}(t)=e^{\mathbf{A}(t-t_0)}\mathbf{h}(t_0)+\int_{t_0}^{t}e^{\mathbf{A}(t-\tau)}\mathbf{B}x(\tau)\mathrm{d}\tau.$$

*Proof.* It is important to note that the proof of Theorem 1 can be found in modern control theory or linear system textbooks. Frist, let us multiply both sides of Eq. $(\mathrm{A}1a)$ by $e^{-\mathbf{A}t}$:

$$e^{-\mathbf{A}t}\dot{\mathbf{h}}(t)=e^{-\mathbf{A}t}\mathbf{A}\mathbf{h}(t)+e^{-\mathbf{A}t}\mathbf{B}x(t), \tag{A25}$$

Then,

$$e^{-\mathbf{A}t}\dot{\mathbf{h}}(t)+e^{-\mathbf{A}t}(-\mathbf{A})\mathbf{h}(t)=e^{-\mathbf{A}t}\mathbf{B}x(t), \tag{A26}$$

Recalling Lemma 1, we have $e^{-\mathbf{A}t}(-\mathbf{A})=(-\mathbf{A})e^{-\mathbf{A}t}$. Thus,

$$e^{-\mathbf{A}t}\dot{\mathbf{h}}(t) + (-\mathbf{A})e^{-\mathbf{A}t}\mathbf{h}(t) = e^{-\mathbf{A}t}\mathbf{B}x(t), \tag{A27}$$

We have known

$$\begin{aligned} \frac{\mathrm{d}\big(e^{-\mathbf{A}t}\mathbf{h}(t)\big)}{\mathrm{d}t} &= e^{-\mathbf{A}t}\dot{\mathbf{h}}(t) + \frac{\mathrm{d}(e^{-\mathbf{A}t})}{\mathrm{d}t}\mathbf{h}(t) \\ &= e^{-\mathbf{A}t}\dot{\mathbf{h}}(t) + (-\mathbf{A})e^{-\mathbf{A}t}\mathbf{h}(t). \end{aligned} \tag{A28}$$

Plug Eq. (A28) into Eq. (A27):

$$\frac{\mathrm{d}\big(e^{-\mathbf{A}t}\mathbf{h}(t)\big)}{\mathrm{d}t} = e^{-\mathbf{A}t}\mathbf{B}x(t). \tag{A29}$$

Therefore,

$$\int_{t_0}^{t} \frac{\mathrm{d}\big(e^{-\mathbf{A}\tau}\mathbf{h}(\tau)\big)}{\mathrm{d}\tau}\,\mathrm{d}\tau = \int_{t_0}^{t} e^{-\mathbf{A}\tau}\mathbf{B}x(\tau)\,\mathrm{d}\tau. \tag{A30}$$

Further,

$$\begin{aligned} e^{-\mathbf{A}\tau}\mathbf{h}(\tau)\Big|_{t_0}^{t} &= e^{-\mathbf{A}t}\mathbf{h}(t) - e^{-\mathbf{A}t_0}\mathbf{h}(t_0) \\ &= \int_{t_0}^{t} e^{-\mathbf{A}\tau}\mathbf{B}x(\tau)\,\mathrm{d}\tau. \end{aligned} \tag{A31}$$

Transposition of terms:

$$e^{-\mathbf{A}t}\mathbf{h}(t) = e^{-\mathbf{A}t_0}\mathbf{h}(t_0) + \int_{t_0}^{t} e^{-\mathbf{A}\tau}\mathbf{B}x(\tau)\,\mathrm{d}\tau. \tag{A32}$$

Therefore,

$$\begin{aligned} \mathbf{h}(t) &= e^{\mathbf{A}t}\left[e^{-\mathbf{A}t_0}\mathbf{h}(t_0) + \int_{t_0}^{t} e^{-\mathbf{A}\tau}\mathbf{B}x(\tau)\,\mathrm{d}\tau\right] \\ &= e^{\mathbf{A}(t-t_0)}\mathbf{h}(t_0) + \int_{t_0}^{t} e^{\mathbf{A}(t-\tau)}\mathbf{B}x(\tau)\,\mathrm{d}\tau. \end{aligned} \tag{A33}$$

Thus, the proof of Theorem 1 is completed. □

Then, let us introduce the ZOH discretization method. Based on Theorem 1, we let $t_0 = t_k$, $t = t_{k+1}$, and $\Delta = t_{k+1} - t_k$. It is worth noting that $t_{k+1} > t_k$ and $\Delta \in \mathbb{R}^{+}$. Then, the solution of the ODE in Theorem 1 can be transformed into

$$\begin{aligned} \mathbf{h}(t_{k+1}) &= e^{\mathbf{A}(t_{k+1}-t_k)}\mathbf{h}(t_k) + \int_{t_k}^{t_{k+1}} e^{\mathbf{A}(t_{k+1}-\tau)}\mathbf{B}x(\tau)d\tau \\ &= e^{\mathbf{A}\Delta}\mathbf{h}(t_k) + \int_{t_k}^{t_{k+1}} e^{\mathbf{A}(t_{k+1}-\tau)}x(\tau)\,d\tau\mathbf{B}. \end{aligned} \tag{A34}$$

In the domain of ZOH, a key assumption is that for $\forall\tau \in [t_k, t_{k+1})$, $x(\tau) = x(t_k)$. Therefore, the above equation can be simplified as

$$\mathbf{h}(t_{k+1}) = e^{\mathbf{A}\Delta}\mathbf{h}(t_k) + \int_{t_k}^{t_{k+1}} e^{\mathbf{A}(t_{k+1}-\tau)}d\tau\mathbf{B}x(t_k). \tag{A35}$$

Next, let us derive the integration $\int_{t_k}^{t_{k+1}} e^{\mathbf{A}(t_{k+1}-\tau)}\,d\tau$. Let $t_{k+1} - \tau = z$, thus we have $d\tau = -dz$. Bring the insight into the integration:

$$\int_{t_k}^{t_{k+1}} e^{\mathbf{A}(t_{k+1}-\tau)}\,d\tau = -\int_{\Delta}^{0} e^{\mathbf{A}z}\,dz$$

$$
\begin{aligned}
&= \int_0^{\Delta} e^{\mathbf{A}z}\, dz \\
&= \mathbf{A}^{-1}(e^{\mathbf{A}\Delta} - \mathbf{I}).
\end{aligned} \tag{A36}
$$

Plug the above the equation into Eq. (A3):

$$\mathbf{h}(t_{k+1}) = e^{\mathbf{A}\Delta}\mathbf{h}(t_k) + \mathbf{A}^{-1}(e^{\mathbf{A}\Delta} - \mathbf{I})\mathbf{B}x(t_k). \tag{A37}$$

We utilize $\mathbf{h}_k$ to denoted $\mathbf{h}(t_k)$ and $x_k$ to represent $x(t_k)$, then the above equation can be changed into

$$\mathbf{h}_{k+1} = e^{\mathbf{A}\Delta}\mathbf{h}_k + \mathbf{A}^{-1}(e^{\mathbf{A}\Delta} - \mathbf{I})\mathbf{B}x(t_k). \tag{A38}$$

Further, it can be transformed into

$$\mathbf{h}_{k+1} = e^{\Delta\mathbf{A}}\mathbf{h}_k + (\Delta\mathbf{A})^{-1}(e^{\Delta\mathbf{A}} - \mathbf{I}) \cdot \Delta\mathbf{B}x_k. \tag{A39}$$

Therefore, recalling Eqs. $(3a)$ and $(3b)$, we have $\overline{\mathbf{A}} = \exp(\Delta\mathbf{A})$, $\overline{\mathbf{B}} = (\Delta\mathbf{A})^{-1}(\exp(\Delta\mathbf{A}) - \mathbf{I}) \cdot \Delta\mathbf{B}$, and $\bar{\mathbf{C}} = \mathbf{C}$.

### A.4 Practical discretization method in Mamba

In the official code of Mamba, a discretization method that is easier to implement compared to the ZOH discretization is adopted to facilitate computation without affecting empirical performance:

$$\overline{\mathbf{A}} = \exp(\Delta\mathbf{A}), \tag{A40$a$}$$

$$\overline{\mathbf{B}} = \Delta\mathbf{B}, \tag{A40$b$}$$

and $\bar{\mathbf{C}} = \mathbf{C}$.

Recalling Eq. (A39), we can see that $\overline{\mathbf{A}}$ here comes from the ZOH approximation. Recall Eq. (A8), it is clear that $\overline{\mathbf{B}}$ here comes from the Euler approximation. Therefore, essentially, the practical discretization method can be viewed as a mix of the ZOH discretization for $\overline{\mathbf{A}}$ and the Euler discretization for $\overline{\mathbf{B}}$ [**].

Besides, the practical discretization method can also be viewed as the first-order approximation of the ZOH discretization method, which is detailed below.

According to Lemma 1, we have derived the Taylor series expansion for the matrix product of $t$:

$$e^{\mathbf{A}t} = \mathbf{I} + \frac{\mathbf{A}t}{1!} + \frac{(\mathbf{A}t)^2}{2!} + \frac{(\mathbf{A}t)^3}{3!} + \cdots. \tag{A41}$$

Further, we change $t$ to the step size $\Delta$ and only consider the zero-order term and the first-order term of the series:

$$e^{\mathbf{A}\Delta} \approx \mathbf{I} + \frac{\mathbf{A}\Delta}{1!} = \mathbf{I} + \mathbf{A}\Delta. \tag{A42}$$

The above formulation is named the first order approximation of $e^{\mathbf{A}\Delta}$. Plug Eq. (A40) into (A37):

$$
\begin{aligned}
\mathbf{h}_{k+1} &= e^{\mathbf{A}\Delta}\mathbf{h}_k + \mathbf{A}^{-1}(e^{\mathbf{A}\Delta} - \mathbf{I})\mathbf{B}x(t_k) \\
&\approx e^{\mathbf{A}\Delta}\mathbf{h}_k + \mathbf{A}^{-1}\mathbf{A}\Delta\mathbf{B}x(t_k) \\
&= e^{\mathbf{A}\Delta}\mathbf{h}_k + \Delta\mathbf{B}x(t_k).
\end{aligned} \tag{A43}
$$

Therefore, $\overline{\mathbf{A}} = \exp(\Delta\mathbf{A})$ and $\overline{\mathbf{B}} = \Delta\mathbf{B}$. The above computations illustrate why the practical discretization method can be viewed as the first-order approximation of the ZOH discretization.

## Appendix B Details of the combination process

Let's restate the discrete state space equations Eqs. $(3a)$ and $(3b)$ for convenience:

$$\mathbf{h}_k = \overline{\mathbf{A}}\mathbf{h}_{k-1} + \overline{\mathbf{B}}x_k, \tag{B1a}$$

$$y_k = \bar{\mathbf{C}}\mathbf{h}_k, \tag{B1b}$$

First, let $k = 0$, and let $\mathbf{h}_{-1} = \mathbf{0}$ for simplicity, we have

[**] https://github.com/state-spaces/mamba/issues/114

$$\mathbf{h}_0 = \overline{\mathbf{B}}x_0, \tag{B2a}$$
$$y_0 = \bar{\mathbf{C}}\mathbf{h}_0 = \bar{\mathbf{C}}\overline{\mathbf{B}}x_0. \tag{B2b}$$

Then, let $k = 1$, and Eqs. $(3a)$ and $(3b)$ yields

$$\mathbf{h}_1 = \overline{\mathbf{A}}\mathbf{h}_0 + \overline{\mathbf{B}}x_1 = \overline{\mathbf{A}}\overline{\mathbf{B}}x_0 + \overline{\mathbf{B}}x_1, \qquad \triangleright \text{ Plug Eq. (B2a)} \tag{B3a}$$
$$y_1 = \bar{\mathbf{C}}\mathbf{h}_1 = \bar{\mathbf{C}}\overline{\mathbf{A}}\overline{\mathbf{B}}x_0 + \bar{\mathbf{C}}\overline{\mathbf{B}}x_1. \tag{B3b}$$

Further, we let $k = 2$ and we can obtain

$$\mathbf{h}_2 = \overline{\mathbf{A}}\mathbf{h}_1 + \overline{\mathbf{B}}x_2 = \overline{\mathbf{A}}\overline{\mathbf{A}}\overline{\mathbf{B}}x_0 + \overline{\mathbf{A}}\overline{\mathbf{B}}x_1 + \overline{\mathbf{B}}x_2 = \overline{\mathbf{A}}^2\overline{\mathbf{B}}x_0 + \overline{\mathbf{A}}\overline{\mathbf{B}}x_1 + \overline{\mathbf{B}}x_2, \qquad \triangleright \text{ Plug Eq. (B3a)} \tag{B4a}$$
$$y_2 = \bar{\mathbf{C}}\mathbf{h}_2 = \bar{\mathbf{C}}\overline{\mathbf{A}}^2\overline{\mathbf{B}}x_0 + \bar{\mathbf{C}}\overline{\mathbf{A}}\overline{\mathbf{B}}x_1 + \bar{\mathbf{C}}\overline{\mathbf{B}}x_2. \tag{B4b}$$

Therefore, we can get the formulations of $\mathbf{h}_k$ and $y_k$ for $\forall k \in \{0,1, \dots, L-1\}$:

$$\mathbf{h}_k = \overline{\mathbf{A}}\mathbf{h}_{k-1} + \overline{\mathbf{B}}x_k = \overline{\mathbf{A}}^k\overline{\mathbf{B}}x_0 + \overline{\mathbf{A}}^{k-1}\overline{\mathbf{B}}x_1 + \cdots + \overline{\mathbf{A}}\overline{\mathbf{B}}x_{k-1} + \overline{\mathbf{B}}x_k, \tag{B5a}$$

$$\begin{aligned} y_k &= \bar{\mathbf{C}}\mathbf{h}_k \\ &= \bar{\mathbf{C}}\overline{\mathbf{A}}^k\overline{\mathbf{B}}x_0 + \bar{\mathbf{C}}\overline{\mathbf{A}}^{k-1}\overline{\mathbf{B}}x_1 + \cdots + \bar{\mathbf{C}}\overline{\mathbf{A}}\overline{\mathbf{B}}x_{k-1} + \bar{\mathbf{C}}\overline{\mathbf{B}}x_k \\ &= \bar{\mathbf{C}}\overline{\mathbf{B}}x_k + \bar{\mathbf{C}}\overline{\mathbf{A}}\overline{\mathbf{B}}x_{k-1} + \cdots + \bar{\mathbf{C}}\overline{\mathbf{A}}^{k-1}\overline{\mathbf{B}}x_1 + \bar{\mathbf{C}}\overline{\mathbf{A}}^k\overline{\mathbf{B}}x_0 \\ &= \sum_{i=0}^{k} \bar{\mathbf{C}}\overline{\mathbf{A}}^i\overline{\mathbf{B}}x_{k-i}. \end{aligned} \tag{B5b}$$

If we let a new sequence be $\overline{\mathbf{K}} = \left(\bar{\mathbf{C}}\overline{\mathbf{B}}, \bar{\mathbf{C}}\overline{\mathbf{A}}\overline{\mathbf{B}}, \dots, \bar{\mathbf{C}}\overline{\mathbf{A}}^{L-1}\overline{\mathbf{B}}\right)$, then Eq. (B5b) can be transferred into

$$y_k = \sum_{i=0}^{k} \overline{\mathbf{K}}_i x_{k-i}. \tag{B6}$$

Then, let's give the definition of the discrete convolution:

**Definition 1.** For functions $f$ and $g$, the discrete convolution of $f$ and $g$ is given by

$$(f \circledast g)[n] = \sum_{m=-\infty}^{\infty} f[m]g[n-m],$$

where $n$ and $m$ are the indexes.

Combining Definition 1 and Eq. (B6), we can obtain

$$\mathbf{y} = \overline{\mathbf{K}} \circledast \mathbf{x}, \tag{B7}$$

where $\overline{\mathbf{K}} \in \mathbb{R}^L$ is the 1D convolutional kernel or 1D convolutional filter.

## Appendix C Explanations of parallel training

### C.1 Parallel training of S4 models

First, let's revisit the parallel training of CNNs. Essentially, a CNN conducts convolution operation between its convolution kernel and the input feature map. Recalling Definition 1, single step of convolution is actually equivalent to a dot product between the convolution kernel and the corresponding size input vector. Because the convolution kernel is fixed during the entire convolution process, CNNs can first save the convolution kernel and then do each step of the convolution operation in parallel.

Because the convolution kernel $\overline{\mathbf{K}}$ is not changed during the entire convolution process, namely, $\mathbf{x} \mapsto \mathbf{y}$, in the training stage, similar to CNNs, S4 models first precompute and cache $\overline{\mathbf{K}}$ and then convolve $\overline{\mathbf{K}}$ with the input $\mathbf{x}$ in parallel to obtain $\mathbf{y}$.

### C.2 The reason why Mamba cannot be converted into the convolutional form

The dimensions of the input sequence $\mathbf{x}$ in the practical problem is $\mathbb{R}^{B \times L \times D}$, where $B$ is the batch size, $L$ denotes the length of the sequence, and $D$ is the dimension of each element in $\mathbf{x}$. Then, Mamba adopted the linear layers to map the dimension $D$ into dimension $N$. Recall Eq. $(11)$:

$$\mathbf{B} = S_{\mathbf{B}}(\mathbf{x}) = \mathrm{Linear}_N(\mathbf{x}) = \mathbf{x} \times \mathbf{W_B} + Bias_{\mathbf{B}} \tag{C1}$$

where $\mathbf{W_B} \in \mathbb{R}^{D\times N}$ and $Bias_{\mathbf{B}} \in \mathbb{R}^{N\times 1}$. Therefore, the shape of $\mathbf{B}$ is $\mathbb{R}^{B\times L\times N}$. Similarly, the shapes of $\mathbf{C}$ and $\Delta$ are $\mathbb{R}^{B\times L\times N}$ and $\mathbb{R}^{B\times L\times D}$, respectively. Here, the new dimensions $B \times L$ among $\mathbf{B}$, $\mathbf{C}$, and $\Delta$ ensure they are input-dependent.

Then, we apply the practical discretization method to discretize $\mathbf{A}$ and $\mathbf{B}$:

$$\overline{\mathbf{A}} = \exp(\Delta\mathbf{A}), \tag{C2$a$}$$

$$\overline{\mathbf{B}} = \Delta\mathbf{B}. \tag{C2$b$}$$

To ensure $\overline{\mathbf{A}}$ and $\overline{\mathbf{B}}$ dynamic, we need to make the dimensions of $\overline{\mathbf{A}}$ and $\overline{\mathbf{B}}$ still contain $B \times L$. Therefore, actually, the multiplication here is not the traditional matrix multiplication anymore. Refreing to the Mamba paper, the shapes of $\overline{\mathbf{A}}$ and $\overline{\mathbf{B}}$ here are both $\mathbb{R}^{B\times L\times D\times N}$.

Next, recall the discretized state space equations in Subsection 2.1:

$$\mathbf{h}_k = \overline{\mathbf{A}}\mathbf{h}_{k-1} + \overline{\mathbf{B}}x_k, \tag{C3a}$$

$$y_k = \bar{\mathbf{C}}\mathbf{h}_k. \tag{C3b}$$

Because the discretized matrices are input-dependent, the forms of Eqs. (C3a) and (C3b) are transformed into:

$$\mathbf{h}_k^b = \overline{\mathbf{A}}_k^b\mathbf{h}_{k-1}^b + \overline{\mathbf{B}}_k^b\mathbf{x}_k^b, \tag{C4a}$$

$$\mathbf{y}_k^b = \bar{\mathbf{C}}_k^b\mathbf{h}_k^b. \tag{C4b}$$

where $b \in \{1,2,\dots,B\}$ denotes the $b$-th sample in a mini batch.

Therefore, for $k = 0$, we have

$$\mathbf{h}_0^b = \overline{\mathbf{B}}_0^b\mathbf{x}_0^b, \tag{C5a}$$

$$\mathbf{y}_0^b = \bar{\mathbf{C}}_0^b\mathbf{h}_0^b = \bar{\mathbf{C}}_0^b\overline{\mathbf{B}}_0^b\mathbf{x}_0^b. \tag{C5b}$$

If we change it into the convolutional form described in Eq. (B6), the convolutional kernel is

$$\overline{\mathbf{K}} = (\bar{\mathbf{C}}_0^b\overline{\mathbf{B}}_0^b, \overline{\mathbf{K}}_1, \overline{\mathbf{K}}_2, \dots, \overline{\mathbf{K}}_{L-1}), \tag{C6}$$

where $\overline{\mathbf{K}}_1, \overline{\mathbf{K}}_2, \dots, \overline{\mathbf{K}}_{L-1}$ represents they are undefined and can be any data. Then, let $k = 1$, and we obtain

$$\mathbf{h}_1^b = \overline{\mathbf{A}}_1^b\mathbf{h}_0^b + \overline{\mathbf{B}}_1^b\mathbf{x}_1^b = \overline{\mathbf{A}}_1^b\overline{\mathbf{B}}_0^b\mathbf{x}_0^b + \overline{\mathbf{B}}_1^b\mathbf{x}_1^b, \tag{C7a}$$

$$\mathbf{y}_1^b = \bar{\mathbf{C}}_1^b\mathbf{h}_1^b = \bar{\mathbf{C}}_1^b\overline{\mathbf{A}}_1^b\overline{\mathbf{B}}_0^b\mathbf{x}_0^b + \bar{\mathbf{C}}_1^b\overline{\mathbf{B}}_1^b\mathbf{x}_1^b. \tag{C7b}$$

The convolutional kernel is

$$\overline{\mathbf{K}} = \left(\bar{\mathbf{C}}_1^b\overline{\mathbf{B}}_1^b, \bar{\mathbf{C}}_1^b\overline{\mathbf{A}}_1^b\overline{\mathbf{B}}_0^b\mathbf{x}_0^b, \overline{\mathbf{K}}_2, \dots, \overline{\mathbf{K}}_{L-1}\right). \tag{C8}$$

Finally, let $k = 2$, and we get

$$\mathbf{h}_2^b = \overline{\mathbf{A}}_2^b\mathbf{h}_1^b + \overline{\mathbf{B}}_2^b\mathbf{x}_2^b = \overline{\mathbf{A}}_2^b\overline{\mathbf{A}}_1^b\overline{\mathbf{B}}_0^b\mathbf{x}_0^b + \overline{\mathbf{A}}_2^b\overline{\mathbf{B}}_1^b\mathbf{x}_1^b + \overline{\mathbf{B}}_2^b\mathbf{x}_2^b, \tag{C9a}$$

$$\mathbf{y}_2^b = \bar{\mathbf{C}}_2^b\mathbf{h}_2^b = \bar{\mathbf{C}}_2^b\overline{\mathbf{A}}_2^b\overline{\mathbf{A}}_1^b\overline{\mathbf{B}}_0^b\mathbf{x}_0^b + \bar{\mathbf{C}}_2^b\overline{\mathbf{A}}_2^b\overline{\mathbf{B}}_1^b\mathbf{x}_1^b + \bar{\mathbf{C}}_2^b\overline{\mathbf{B}}_2^b\mathbf{x}_2^b. \tag{C9b}$$

At this time, the convolutional kernel is

$$\overline{\mathbf{K}} = \left(\bar{\mathbf{C}}_2^b\overline{\mathbf{B}}_2^b, \bar{\mathbf{C}}_2^b\overline{\mathbf{A}}_2^b\overline{\mathbf{B}}_1^b, \bar{\mathbf{C}}_2^b\overline{\mathbf{A}}_2^b\overline{\mathbf{A}}_1^b\overline{\mathbf{B}}_0^b, \overline{\mathbf{K}}_3 \dots, \overline{\mathbf{K}}_{L-1}\right). \tag{C10}$$

Based on Eqs. (C6), (C8), and (C10), we can find that the convolutional kernels for different indexes of the output sequence are not unified. Therefore, we cannot utilize the convolutional form to represent the state space equations anymore.

## Appendix D Explanations of the complexity of S4 models

First, it is clear that the shape of $\overline{\mathbf{K}}_k$ is $\mathbb{R}^{N\times L}$, therefore, the space complexity is $O(NL)$. Then, let's talk about the computational cost. Assuming that we have computed $\overline{\mathbf{A}}^k\overline{\mathbf{B}} \in \mathbb{R}^{N\times 1}$ (recall $\overline{\mathbf{A}} \in \mathbb{R}^{N\times N}$, and $\overline{\mathbf{B}} \in \mathbb{R}^{N\times 1}$), we need to obtain $\overline{\mathbf{A}}^{k+1}\overline{\mathbf{B}}$ next step, and it can be computed by matrix-vector multiplication (MVM):

$$\overline{\mathbf{A}}^{k+1}\overline{\mathbf{B}} = \overline{\mathbf{A}} \times \left(\overline{\mathbf{A}}^k\overline{\mathbf{B}}\right). \tag{D1}$$

Therefore, for the single stet, the computation process involves $N \times N$ multiplication operations. Because the total calculation of $\overline{\mathbf{K}}_k$ needs $L$ successive multiplications by $\overline{\mathbf{A}}$, the computational cost is equal to $N^2L$, i.e., the computation complexity is $O(N^2L)$.

## Appendix E Effective receptive fields (ERFs)

### E.1 Receptive field (RF) and ERF

Based on the description in [51], we can first give the definition of the receptive field (RF) or the field of view:

**Definition 2** (Receptive field (RF))**.** For a neural network, let the input image be $\mathbf{I} \in \mathbb{R}^{C\times H\times W}$, and the output feature map be $\mathbf{M} \in \mathbb{R}^{C^O\times H^O\times W^O}$, where $C$ and $C^O$ denote the channels of the input image and the output map, and $(H \times W)$ and $(H^O \times W^O)$ represent the spatial resolutions, where $H$ and $W$ mean the height and width, respectively. For any unit $u \in \mathbf{M}$, it values only depends on a certain region $\mathbf{R} \subset \mathbf{I}$. The region in the input images is defined as the RF of the unit.

**Remark 1.** It is clear that different pixels in the region $\mathbf{R}$ have different impacts on the value of $u$. In other words, the contributions of the units in $\mathbf{R}$ to the value of $u$ are different. Intuitively, let's consider a unit $u_1 = M_{c_1,h_1,w_1}$. The set $\{I_{c,h_1,w_1}|1 \le c \le C\}$ can always bring a larger impact to $u_1$ than other units in $\mathbf{I}$ since its information can be sent to $u_1$ by multiple paths. In order to mathematically and quantitatively represent the contributions, Luo et al. [51] propose the effective receptive field (ERF). The definition of ERF is given as follows:

**Definition 3** (Effective receptive field (ERF))**.** For simplicity, ERF only considers the impacts of the units in $\mathbf{I}$ on the central units in $\mathbf{M}$. The set of the central units can be denoted as $U = \{M_{c,H^O/2,W^O/2}|1 \le c \le C^O\}$. For one central unit $u_j = M_{j,H^O/2,W^O/2}, 1 \le j \le C^O$ belongs to $U$, the contribution value of any pixel $I_{c,h,w}$ in $\mathbf{I}$ to $u_j$ can be defined as the partial derivative $\partial u_j/\partial I_{c,h,w}$, which measures how much $u_j$ changes as $I_{c,h,w}$ changes by a small amount. Further, the ERF of $u_j$ is defined as the set of units in $\mathbf{I}$ that have a non-negligible impact on $u_j$. If we set the threshold value of the contribution to be $T$, the ERF of $u_j$, $E$, is

$$E = \left\{ I_{c,h,w} \in \mathbf{I} \middle| \frac{\partial u_j}{\partial I_{c,h,w}} \ge T \right\}.$$

**Remark 2.** It is clear that the contribution $\partial u_j/\partial I_{c,h,w}$ not only depends on the weights and biases of the neural network but also is input-dependent based on the knowledge of back-propagation.

### E.2 Visualization of ERFs

**Definition 4** (Contribution map)**.** Based on Definition 3, we can obtain a contribution matrix $\mathbf{N}^j \in \mathbb{R}^{C\times H\times W}$ for each $u_j$. Each unit in $\mathbf{N}^j$ is the contribution value of the corresponding pixel in $\mathbf{I}$ to $u_j$. Recalling Definition 3, $\mathbf{N}^j$ can be formulated as

$$N^j_{c,h,w} = \frac{\partial u_j}{\partial I_{c,h,w}}, 1 \le j \le C^O. \tag{E1}$$

**Remark 4.** Based on the definition of the contribution matrix, it is clear that the shape of the contribution map is the same as that of $\mathbf{I}$.

**Definition 5** (Visualization image of ERF)**.** The visualization image of ERF is a single channel contribution map obtained by compressing the contribution matrices.

Recalling Remark 2, the contribution values are input-dependent, and thus the contribution matrix is also input-dependent. We aim for the visualized image of ERF to reflect the overall and comprehensive performance of the contribution matrices from different inputs. Therefore, in practice, $\mathbf{I}$ and $\mathbf{M}$ are extended to four dimensions with the first dimension denoting the input images.

Ding et al. [47] propose a practical approach to generate the visualization image of ERF, which is summarized as follows. First, they consider $B$ input images, thus, the new shapes of $\mathbf{I}$ and $\mathbf{M}$ can be defined as $\mathbf{I} \in \mathbb{R}^{B\times C\times H\times W}$ and $\mathbf{M} \in \mathbb{R}^{B\times C^O\times H^O\times W^O}$, respectively. At this time, for the new output feature map $\mathbf{M}$, there are $B \times C^O$ central points, $u_{b,j}$, $1 \le b \le B, 1 \le j \le C^O$ and the corresponding $B \times C^O$ contribution maps $\mathbf{N}^{b,j} \in \mathbb{R}^{B\times C\times H\times W}$ (recall Remark 4). Then, they calculate an overall contribution matrix $\mathbf{ON} \in \mathbb{R}^{B\times C\times H\times W}$ for all the $B \times C^O$ central points:

$$ON_{b_1,c_1,h_1,w_1} = \sum_{b=1}^{B}\sum_{j=1}^{C^O} N^{b,j}_{b_1,c_1,h_1,w_1}$$

$$= \sum_{b=1}^{B}\sum_{j=1}^{C^O} \frac{\partial u_{b,j}}{\partial I_{b_1,c_1,h_1,w_1}}. \tag{E2}$$

Next, Ding et al. [47] set the threshold value $T$ to be 0. Thus, if there is a pixel whose contribution value to a central unit is less than 0, we can consider it to have no impact on the central point, meaning the pixel does not belong to the ERF of the central point (refer to Definition 3). Therefore, we can change the contribution value of the pixel to 0, indicating there is no impact. After removing the negative parts, $\mathbf{ON}$ is converted to the matrix $\mathbf{P} \in \mathbb{R}^{B\times C\times H\times W}$:

$$P_{b,c,h,w} = \max(ON_{b,c,h,w}, 0)\,. \tag{E3}$$

Next, they aggregate all the units in across all the examples and input channels (namely, add all the values along $B$ and $C$) to obtain the single channel contribution map $\mathbf{A} \in \mathbb{R}^{H\times W}$:

$$A_{h,w} = \log_{10}\left(\sum_{b=1}^{B}\sum_{c=1}^{C} P_{b,c,h,w} + 1\right). \tag{E3}$$

where they add 1 to make $A_{h,w} > 0$. Last, the values of $\mathbf{A}$ are rescaled to [0,1] by dividing the maximum value in $\mathbf{A}$ for comparability across all the models:

$$\mathbf{A} = \frac{\mathbf{A}}{\max(A_{h,w})}. \tag{E3}$$

Based on the following calculations, we can see that (1) the shape of $\mathbf{A}$ is the same as the spatial resolution of the input image $\mathbf{I}$; (2) each value in $\mathbf{A}$ represents the overall contribution or impact of the corresponding spatial pixel in $\mathbf{I}$ to the central point (in the spatial dimensions) in $\mathbf{M}$; and (3) the higher the value in $\mathbf{A}$, the greater the contribution of the spatial pixel in $\mathbf{I}$ to the central point, namely, the stronger the perception of the central point to the pixel.

## References

[1] W. Chen, Z. He, J. Zhang, Online monitoring of crack dynamic development using attention-based deep networks, Automation in Construction. 154 (2023), 105022, https://doi.org/10.1016/j.autcon.2023.105022.

[2] Z. He, W. Chen, J. Zhang, Y.-H. Wang, Crack segmentation on steel structures using boundary guidance model, Automation in Construction. 162 (2024), 105354, https://doi.org/10.1016/j.autcon.2024.105354.

[3] Y.-J. Cha, W. Choi, O. Büyüköztürk, Deep learning-based crack damage detection using convolutional neural networks, Computer-Aided Civil and Infrastructure Engineering. 32 (5) (2017), pp. 361-378, https://doi.org/10.1111/mice.12263.

[4] J. Chow, Z. Su, J. Wu, P. Tan, X. Mao, Y.-H. Wang, Anomaly detection of defects on concrete structures with the convolutional Autoencoder, Advanced Engineering Informatics. 45 (2020), 101105, https://doi.org/10.1016/j.aei.2020.101105.

[5] J. Guo, P. Liu, B. Xiao, L. Deng, Q. Wang, Surface defect detection of civil structures using images: Review from data perspective, Automation in Construction. 158 (2024), 105186, https://doi.org/10.1016/j.autcon.2023.105186.

[6] Y. Xu, Y. Bao, J. Chen, W. Zuo, H. Li, Surface fatigue crack identification in steel box girder of bridges by a deep fusion convolutional neural network based on consumer-grade camera images, Structural Health Monitoring. 18 (3) (2019), pp. 653-674, https://doi.org/10.1177/1475921718764873.

[7] F. Ni, Z. He, S. Jiang, W. Wang, J. Zhang, A Generative adversarial learning strategy for enhanced lightweight crack delineation networks, Advanced Engineering Informatics. 52 (2022), 101575, https://doi.org/10.1016/j.aei.2022.101575.

[8] A. Vaswani, N. Shazeer, N. Parmar, J. Uszkoreit, L. Jones, A.N. Gomez, Ł. Kaiser, I. Polosukhin, Attention is all you need, in: 2017 International Conference on Neural Information Processing Systems (NeurIPS), ACM, Long Beach, CA, USA, 2017, pp. 6000-6010, https://dl.acm.org/doi/10.5555/3295222.3295349.

[9] Y. Lu, W. Qin, C. Zhou, Z. Liu, Automated detection of dangerous work zone for crawler crane guided by UAV images via Swin Transformer, Automation in Construction. 147 (2023), 104744, https://doi.org/10.1016/j.autcon.2023.104744.

[10] T. Gao, Z. Yuanzhou, B. Ji, Z. Xie, Vision-based fatigue crack automatic perception and geometric updating of finite element model for welded joint in steel structures, Computer-Aided Civil and Infrastructure Engineering. 39 (11) (2024), pp. 1659-1675, https://doi.org/10.1111/mice.13166.

[11] S. Mehta, M. Rastegari, MobileViT: Lightweight, general-purpose, and mobile-friendly vision transformer, in: 2022 International Conference on Learning Representations (ICLR), Online, 2022, pp.1-26, https://openreview.net/forum?id=vh-0sUt8HlG.

[12] Z. Liu, Y. Lin, Y. Cao, H. Hu, Y. Wei, Z. Zhang, S. Lin, B. Guo, Swin transformer: Hierarchical vision transformer using shifted windows, in: 2021 IEEE/CVF International Conference on Computer Vision (ICCV), IEEE, Montreal, QC, Canada, 2021, pp. 9992-10002, https://doi.org/10.1109/ICCV48922.2021.00986.

[13] R. Child, S. Gray, A. Radford, I. Sutskever, Generating long sequences with sparse transformers, arXiv:1904.10509v1, https://arxiv.org/abs/1904.10509, 2019.

[14] A. Katharopoulos, A. Vyas, N. Pappas, F. Fleuret, Transformers are RNNs: Fast autoregressive transformers with linear attention, in: 37th International Conference on Machine Learning (ICML), PMLR, Online, 2020, pp. 5156-5165, https://proceedings.mlr.press/v119/katharopoulos20a.html.

[15] T. Dao, D.Y. Fu, S. Ermon, A. Rudra, C. Ré, FlashAttention: Fast and memory-efficient exact attention with IO-Awareness, arXiv:2205.14135v2, https://arxiv.org/abs/2205.14135, 2022.

[16] A. Gu, T. Dao, Mamba: Linear-time sequence modeling with selective state spaces, arXiv:2312.00752v2, https://arxiv.org/abs/2312.00752, 2024.

[17] A. Gu, I. Johnson, K. Goel, K. Saab, T. Dao, A. Rudra, C. Re. Combining recurrent, convolutional, and continuous-time models with linear state-space layers, arXiv:2110.13985v1, https://arxiv.org/abs/2110.13985, 2021.

[18] A. Gu, K. Goel, C. Re. Efficiently modeling long sequences with structured state spaces, arXiv:2111.00396v3, https://arxiv.org/abs/2111.00396, 2022.

[19] L. Zhu, B. Liao, Q. Zhang, X. Wang, W. Liu, X. Wang, Vision Mamba: Efficient visual representation learning with bidirectional state space model, arXiv:2401.09417v2, https://arxiv.org/abs/2401.09417, 2024.

[20] Y. Liu, Y. Tian, Y. Zhao, H. Yu, VMamba: Visual state space model, arXiv:2401.10166v2, https://arxiv.org/abs/2401.10166, 2024.

[21] Z. Wang, J.-Q. Zheng, Y. Zhang, G. Cui, L. Li, Mamba-UNet: UNet-like pure visual mamba for medical image segmentation, arXiv:2402.05079v2, https://arxiv.org/abs/2402.05079, 2024.

[22] X. Ma, X. Zhang, M.-O. Pun, RS3Mamba: Visual state space model for remote sensing images semantic segmentation, arXiv:2404.02457v1, https://arxiv.org/abs/2404.02457, 2024.

[23] Z. He, S. Jiang, J. Zhang, G. Wu, Automatic damage detection using anchor-free method and unmanned surface vessel,

Automation in Construction. 133 (2022), 104017, https://doi.org/10.1016/j.autcon.2021.104017.

[24] H. Chu, W. Chen, L. Deng, Cascade operation-enhanced high-resolution representation learning for meticulous segmentation of bridge cracks, Advanced Engineering Informatics. 61 (2024), 102508, https://doi.org/10.1016/j.aei.2024.102508.

[25] Z. He, C. Su, Y. Deng, A novel MO-YOLOv4 for segmentation of multi-class bridge damages, Advanced Engineering Informatics. 62 (2024), 102586, https://doi.org/10.1016/j.aei.2024.102586.

[26] Z. He, Y.-H. Wang, J. Zhang, Generative structural design integrating BIM and diffusion model, arXiv: 2311.04052v1, https://arxiv.org/abs/2311.04052, 2023.

[27] A. Dosovitskiy, L. Beyer, A. Kolesnikov, D. Weissenborn, X. Zhai, T. Unterthiner, M. Dehghani, M. Minderer, G. Heigold, S. Gelly, J. Uszkoreit, N. Houlsby, An image is worth 16x16 words: Transformers for image recognition at scale, in: 2021 International Conference on Learning Representations (ICLR), Online, 2021, pp. 1-21, https://openreview.net/forum?id=YicbFdNTTy.

[28] J. L. Ba, J. R. Kiros, G. E. Hinton, Layer normalization, arXiv:1607.06450, https://arxiv.org/abs/1607.06450, 2016 (accessed 18 January 2022).

[29] J. Hu, L. Shen, S. Albanie, G. Sun, E. Wu, Squeeze-and-excitation networks, IEEE Transactions on Pattern Analysis and Machine Intelligence. 42 (8) (2020), pp. 2011-2023, https://doi.org/10.1109/TPAMI.2019.2913372.

[30] S. Woo, J. Park, J.-Y. Lee, I.S. Kweon, CBAM: Convolutional block attention module, in: 2018 European Conference on Computer Vision (ECCV), Springer, Cham, Munich, Germany, 2018, pp. 3-19, https://doi.org/10.1007/978-3-030-01234-2_1.

[31] L. Zhang, G. Wang, W. Sun, Automatic identification of building structure types using unmanned aerial vehicle oblique images and deep learning considering facade prior knowledge, International Journal of Digital Earth. 16 (1) (2023) pp. 3348-3367, https://doi.org/10.1080/17538947.2023.2247390.

[32] I.O. Agyemang, X. Zhang, I. Adjei-Mensah, D. Acheampong, L.D. Fiasam, C. Sey, S.B. Yussif, D. Effah, Automated vision-based structural health inspection and assessment for post-construction civil infrastructure, Automation in Construction. 156 (2023), 105153, https://doi.org/10.1016/j.autcon.2023.105153.

[33] G.-Q. Zhang, B. Wang, J. Li, Y.-L. Xu, The application of deep learning in bridge health monitoring: a literature review, Advances in Bridge Engineering. 3 (2022), 22, https://doi.org/10.1186/s43251-022-00078-7.

[34] Q. Min, M. Zhang, M. Li, Y. He, S.P.A. Bordas, H. Zhang, Hydraulic fracturing simulation of concrete dam integrating intelligent crack detection and refined modeling methods, Engineering Structures. 305 (2024), 117760, https://doi.org/10.1016/j.engstruct.2024.117760.

[35] O. Ronneberger, P. Fischer, T. Brox, U-Net: Convolutional networks for biomedical image segmentation, in: 2015 International Conference on Medical Image Computing and Computer-Assisted Intervention (MICCAI), Springer, Cham, Munich, Germany, 2015, pp. 234-241, https://doi.org/10.1007/978-3-319-24574-4_28.

[36] Z. Zhou, M.M.R. Siddiquee, N. Tajbakhsh, J. Liang, UNet++: Redesigning skip connections to exploit multiscale features in image segmentation, IEEE Transactions on Medical Imaging. 39 (6) (2020), pp. 1856-1867, https://doi.org/10.1109/TMI.2019.2959609.

[37] O. Oktay, J. Schlemper, L.L. Folgoc, M. Lee, M. Heinrich, K. Misawa, K. Mori, S. McDonagh, N.Y. Hammerla, B. Kainz, B. Glocker, D. Rueckert, Attention U-Net: Learning where to look for the pancreas, arXiv:1804.03999v3, https://arxiv.org/abs/1804.03999, 2018 (accessed 1 May 2022).

[38] Q. Zou, Y. Cao, Q. Li, Q. Mao, S. Wang, CrackTree: Automatic crack detection from pavement images, Pattern Recognition Letters. 33 (2012), pp. 227-238, https://doi.org/10.1016/j.patrec.2011.11.004.

[39] Y. Shi, L. Cui, Z. Qi, F. Meng, Z. Chen, Automatic road crack detection using random structured forests, IEEE Transactions on Intelligent Transportation Systems. 17 (12) (2016), pp. 3434-3445, https://doi.org/10.1109/TITS.2016.2552248.

[40] F. Yang, L. Zhang, S. Yu, D. Prokhorov, X. Mei, H. Ling, Feature pyramid and hierarchical boosting network for pavement crack

detection, IEEE Transactions on Intelligent Transportation Systems. 21 (4) (2020), pp. 1525-1535, https://doi.org/10.1109/TITS.2019.2910595.

[41] Y. Liu, J. Yao, X. Lu, R. Xie, L. Li, DeepCrack: A deep hierarchical feature learning architecture for crack segmentation, Neurocomputing. 338 (2019), pp. 139-153, https://doi.org/10.1016/j.neucom.2019.01.036.

[42] Z. Gu, J. Cheng, H. Fu, K. Zhou, H. Hao, Y. Zhao, T. Zhang, S. Gao, J. Liu, CE-Net: Context encoder network for 2D medical image segmentation, IEEE Transactions on Medical Imaging. 38 (10) (2019), pp. 2281-2292, https://doi.org/10.1109/TMI.2019.2903562.

[43] L.-C. Chen, Y. Zhu, G. Papandreou, F. Schroff, H. Adam, Encoder-decoder with atrous separable convolution for semantic image segmentation, in: 2018 European Conference on Computer Vision (ECCV), Springer, Cham, Munich, Germany, 2018, pp. 833-851, https://doi.org/10.1007/978-3-030-01234-2_49.

[44] Z. Wu, L. Su, Q. Huang, Stacked cross refinement network for edge-aware salient object detection, in: 2019 IEEE/CVF International Conference on Computer Vision (ICCV), IEEE, Seoul, Korea (South), 2019, pp. 7263-7272, https://doi.org/10.1109/ICCV.2019.00736.

[45] J. Chen, Y. Lu, Q. Yu, X. Luo, E. Adeli, Y. Wang, TransUNet: Transformers make strong encoders for medical image segmentation, arXiv:2102.04306v1, https://arxiv.org/abs/2102.04306, 2021 (accessed 1 October 2022).

[46] Z. Yang, S. Farsiu, Directional connectivity-based segmentation of medical images, in: 2023 IEEE/CVF Conference on Computer Vision and Pattern Recognition (CVPR), IEEE, Vancouver, BC, Canada, 2023, pp. 11525-11535, https://doi.org/10.1109/CVPR52729.2023.01109.

[47] X. Ding, X. Zhang, J. Han, G. Ding, Scaling up your kernels to 31×31: Revisiting large kernel design in cnns, in: 2022 IEEE/CVF Conference on Computer Vision and Pattern Recognition (CVPR), IEEE, New Orleans, LA, USA, 2022, pp. 11953-11965, https://doi.org/10.1109/CVPR52688.2022.01166.

[48] X. Yang, E.R. Castillo, Y. Zou, L. Wotherspoon, UAV-deployed deep learning network for real-time multi-class damage detection using model quantization techniques, Automation in Construction. 159 (2024), 105254, https://doi.org/10.1016/j.autcon.2023.105254.

[49] Z. Chen, D.-C. Feng, X.-Y. Cao, G. Wu, Time-variant seismic resilience of reinforced concrete buildings subjected to spatiotemporal random deterioration, Engineering Structures. 305 (2024). 117759, https://doi.org/10.1016/j.engstruct.2024.117759.

[50] J. Zhang, L. Zhou, Y. Tian, S. Yu, W. Zhao, Y. Cheng, Vortex-induced vibration measurement of a long-span suspension bridge through noncontact sensing strategies, Computer-Aided Civil and Infrastructure Engineering. 37 (12) (2022), pp. 1617-1633, https://doi.org/10.1111/mice.12712.

[51] W. Luo, Y. Li, R. Urtasun, R. Zemel, Understanding the effective receptive field in deep convolutional neural networks, in: 30th International Conference on Neural Information Processing Systems (NIPS), ACM, Barcelona, Spain, 2016, pp. 4905-4913, https://dl.acm.org/doi/10.5555/3157382.3157645.